\documentclass[Afour,sageh,times]{sagej}

\usepackage{moreverb,url}

\usepackage[colorlinks,bookmarksopen,bookmarksnumbered,citecolor=red,urlcolor=red]{hyperref}

\newcommand\BibTeX{{\rmfamily B\kern-.05em \textsc{i\kern-.025em b}\kern-.08em
T\kern-.1667em\lower.7ex\hbox{E}\kern-.125emX}}

\def\volumeyear{2016}

\usepackage{graphics} 
\usepackage{epsfig} 
\usepackage{times} 
\usepackage{amsmath} 
\usepackage{amssymb}  
\usepackage{algorithm}
\usepackage[noend]{algpseudocode}
\usepackage[tableposition=top]{caption}
\usepackage{subcaption}

\usepackage[shortlabels]{enumitem}
\usepackage{multirow}
\usepackage{booktabs}

\usepackage{url,graphicx,tabularx,array,geometry,amsthm,mathrsfs,comment}
\usepackage{makeidx} 

\usepackage{verbatim,epstopdf,bm}
\usepackage{float}
\floatstyle{plaintop}
\restylefloat{table}
\usepackage{pgfplots}
\pgfplotsset{compat=newest}
\usepackage{pgfplotstable}
\pgfplotsset{every axis/.append style={font=\footnotesize}}
\usepackage{titlesec}
\usepackage{filemod}
\usepackage{wrapfig}
\usepackage{textgreek}

\usepackage[utf8]{inputenc}

\usepackage{csquotes}

\usepackage[english]{babel}

\usepackage{hyperref}
\hypersetup{linktocpage}
\hypersetup{colorlinks,
citecolor=blue
}

\usepackage[perpage,hang]{footmisc}

\usepackage[disable]{todonotes}

\algdef{SE}[DOWHILE]{Do}{doWhile}{\algorithmicdo}[1]{\algorithmicwhile\ #1}%
\algrenewcommand\algorithmicindent{1.0em}%

\newif\ifproofread

\newcommand{\changemarker}[1]{%
\ifproofread
\textcolor{black}{#1}%
\else
#1%
\fi
}

\proofreadtrue

\newcommand{\pmnn}{\textsc{PMNN}}
\newcommand{\sensorfeatures}{\boldsymbol{s}}
\newcommand{\sensortraces}{\boldsymbol{s}}
\newcommand{\sensortracesdeviation}{\boldsymbol{\Delta} \sensortraces}

\newcommand{\sensorexp}{\sensortraces_\text{expected}}
\newcommand{\sensoract}{\sensortraces_\text{actual}}

\newcommand{\phasevariable}{p}
\newcommand{\phasevelocity}{u}
\newcommand{\forcingterm}{\boldsymbol{f}}
\newcommand{\forcingtermscalar}{f}
\newcommand{\couplingterm}{\boldsymbol{c}}
\newcommand{\couplingtermscalar}{c}

\newcommand{\quatforcingterm}{\forcingterm}
\newcommand{\quatcouplingterm}{\couplingterm}
\newcommand{\targetforcingterm}{\forcingterm_\text{target}}
\newcommand{\targetcouplingterm}{\couplingterm_\text{target}}

\newcommand{\targetquatforcingterm}{\targetforcingterm}
\newcommand{\targetquatcouplingterm}{\targetcouplingterm}

\newcommand{\normalizeddmpbasisfunc}{\boldsymbol{\Phi}}
\newcommand{\normalizeddmpbasisfuncscalar}{\phi}

\newcommand{\quatstateposition}{\boldsymbol{Q}}
\newcommand{\angularvelocity}{\boldsymbol{\omega}}
\newcommand{\angularacceleration}{\dot{\angularvelocity}}
\newcommand{\quatevolvinggoalstateposition}{\quatstateposition_{g}}
\newcommand{\quatsteadygoalstateposition}{\quatstateposition_{G}}

\newcommand{\dmpmotiondurationprop}{\tau} 

\newcommand{\reidentifier}{\text{ref}}
\newcommand{\targetidentifier}{\text{guess}}
\newcommand{\retraj}{\traj_\reidentifier}
\newcommand{\targettraj}{\traj_\targetidentifier}
\newcommand{\idxbaselinedemo}{l}
\newcommand{\numbaselinedemos}{L}
\newcommand{\unsegmentedtraj}{\boldsymbol{\mathcal{U}}}
\newcommand{\indexedunsegmentedtraj}{\unsegmentedtraj_{\idxbaselinedemo}}
\newcommand{\firstunsegmentedtraj}{\unsegmentedtraj_{1}}
\newcommand{\zvcthresh}{h}
\newcommand{\idxprimitive}{p}
\newcommand{\numprimitives}{P}

\newcommand{\unsegmentedtrajdatapoint}{{z}}

\newcommand{\primretrajstartidx}{s_1^\idxprimitive}
\newcommand{\primretrajendidx}{e_1^\idxprimitive}
\newcommand{\primtargettrajstartidx}{s_\idxbaselinedemo^\idxprimitive}
\newcommand{\primtargettrajendidx}{e_\idxbaselinedemo^\idxprimitive}

\newcommand{\zvcfunction}{ZVC}
\newcommand{\dtwfunction}{DTW}

\newcommand{\correspondencepairslist}{\mathcal{C}}
\newcommand{\correspondencepointre}{t^\reidentifier}
\newcommand{\correspondencepointtarget}{t^\targetidentifier}
\newcommand{\lsconstructfunction}{ConstructLS}
\newcommand{\lswcomputefunction}{ComputeLSWeights}
\newcommand{\wlssolvefunction}{SolveWLS}
\newcommand{\timescale}{\zeta}
\newcommand{\timedelay}{d}

\newcommand{\targettimescale}{\timescale_{\idxbaselinedemo}}
\newcommand{\targettimedelay}{\timedelay_{\idxbaselinedemo}}
\newcommand{\extractednominalprimslist}{\mathcal{S}}
\newcommand{\dtwsearchspaceextension}{\epsilon}

\newcommand{\dmpparam}{{{\theta}_{\forcingtermscalar}}}
\newcommand{\dmpparamset}{\boldsymbol{\dmpparam}}
\newcommand{\sampleindexeddmpparamset}{\dmpparamset_{k}}
\newcommand{\timeindexeddmpparamset}{\dmpparamset_{t}}
\newcommand{\timeindexedpolicyexplorationcovariance}{\policyexplorationcovariance_{t}}

\newcommand{\correcteddmp}{c}
\newcommand{\indexedperturbedcorrecteddmp}{\correcteddmp_{k}}
\newcommand{\baselinedmpparamset}{\dmpparamset}
\newcommand{\correcteddmpparamset}{\dmpparamset_{\correcteddmp}}

\newcommand{\newcorrecteddmpparamset}{\dmpparamset_{\correcteddmp}{'}}
\newcommand{\indexedperturbedcorrecteddmpparamset}{\dmpparamset_{\indexedperturbedcorrecteddmp}}
\newcommand{\pmnnparamset}{\boldsymbol{\theta}_{\pmnn}}

\newcommand{\dataset}{\mathcal{D}}
\newcommand{\correcteddemodataset}{\dataset_\text{cdemo}}
\newcommand{\additionalcorrecteddemodataset}{\dataset_\text{cdemo, additional}}
\newcommand{\fbmodeltrainingdemodataset}{\dataset_\text{fb\_train}}

\newcommand{\DMPtrainfunction}{train\_DMP}
\newcommand{\DMPFBtrainfunction}{train\_DMP\_FB}
\newcommand{\DMPFBunrollfunction}{unroll}
\newcommand{\traj}{\boldsymbol{\mathcal{T}}}
\newcommand{\trajDMPFBunroll}{\traj}
\newcommand{\newproxytrajDMPFBunroll}{\trajDMPFBunroll_\text{new}'}

\newcommand{\zeroparam}{\boldsymbol{0}}

\newcommand{\policyexplorationcovariance}{\boldsymbol{\Sigma}}

\newcommand{\samplingfunction}{sample}
\newcommand{\normaldistribution}{\mathcal{N}}
\newcommand{\costscalar}{J}
\newcommand{\cost}{\boldsymbol{\costscalar}}
\newcommand{\costtogo}{S}
\newcommand{\sampleprobability}{P}
\newcommand{\proxycost}{\cost'}
\newcommand{\indexedcost}{\cost_{k}}
\newcommand{\indexedproxycost}{\proxycost_{k}}

\newcommand{\costthreshold}{\costscalar_\text{thr}}

\newcommand{\pisquaredcmaupdatefunction}{PI^2CMA}

\newcommand{\pisquaredpolicynoise}{\boldsymbol{\epsilon}}

\newcommand\numberthis{\addtocounter{equation}{1}\tag{\theequation}}

\newcommand{\norm}[1]{\left\lVert#1\right\rVert}

\newcommand{\T}{^{\textrm T}} 

\begin{document}

\runninghead{Sutanto et al.}

\title{Supervised Learning and \\Reinforcement Learning of \\Feedback Models for \\Reactive Behaviors: \\Tactile Feedback Testbed}

\author{
Giovanni Sutanto\affilnum{1,2,3,*}, Katharina Rombach\affilnum{4,+}, Yevgen Chebotar\affilnum{5,*}, Zhe Su\affilnum{6,*}, Stefan Schaal\affilnum{3,*}, Gaurav S. Sukhatme\affilnum{2} and Franziska Meier\affilnum{7}
}

\affiliation{
\affilnum{1}Autonomous Motion Department, Max Planck Institute for Intelligent Systems (AMD at MPI-IS), T\"ubingen, Germany\\
\affilnum{2}Department of Computer Science, University of Southern California (CS at USC), Los Angeles, CA, USA\\
\affilnum{3}Intrinsic Innovation LLC., Mountain View, CA, USA\\
\affilnum{4}Chair of Intelligent Maintenance Systems, ETH Zurich, Switzerland\\
\affilnum{5}Google Brain, Mountain View, CA, USA\\
\affilnum{6}Dexterity Inc., Redwood City, CA, USA\\
\affilnum{7}Meta Artificial Intelligence, Menlo Park, CA, USA\\
\affilnum{*}Contributions were made during past affiliations with AMD at MPI-IS and CS at USC.\\
\affilnum{+}Contributions were made during past affiliation with AMD at MPI-IS.
}

\corrauth{Giovanni Sutanto, Univ. of Southern California, Los Angeles, CA, USA.}
\email{gsutanto@alumni.usc.edu}

\begin{abstract}
    \textit{Robots need to be able to adapt to unexpected changes in the environment such that they can autonomously succeed in their tasks. However, hand-designing feedback models for adaptation is tedious, if at all possible, making data-driven methods a promising alternative. In this paper we introduce a full framework for learning feedback models for reactive motion planning. Our pipeline starts by segmenting demonstrations of a complete task into motion primitives via a semi-automated segmentation algorithm. Then, given additional demonstrations of successful adaptation behaviors, we learn initial feedback models through learning from demonstrations. In the final phase, a sample-efficient reinforcement learning algorithm fine-tunes these feedback models for novel task settings through few real system interactions. We evaluate our approach on a real anthropomorphic robot in learning a tactile feedback task.}
\end{abstract}


\maketitle

\section{Introduction}
\label{sec:rlfb_intro}
The ability to deal with changes in the environment is crucial for every robot operating in dynamic environments. 
In order to reliably accomplish its tasks, the robot needs to adapt its plan if unexpected changes arise in the environment.
For instance, a grasping manipulation task involves a sequence of motions such as reaching an object and grasping it. While executing the grasp plan, environment changes may occur, requiring an online modulation of the plan, such as avoiding collisions while reaching the object, or adapting a grasp skill to account for object shape variations.

Adaptation in motion planning can be achieved either by re-planning \citep{Park_ITOMP_ICAPS12, Ratliff_CHOMP, Byravan_TCHOMP, Ratliff_RieMO_ICRA15, Mukadam_GPMP2}
or by reactive motion planning via feedback models \changemarker{--i.e. models which map deviations in sensor space to changes in action--} \citep{Pairet_IROS19, Kober_IROS_2008, Park_Humanoids_2008, Hoffmann_ICRA_2009, pastor_IROS_2011_ASM, icra2017_learning_feedback, icra2018_learn_tactile_feedback}. Online adaptation via re-planning is done via trajectory optimization, which is computationally expensive due to its iterative process that can be slow to converge to a feasible and optimal solution. On the other hand, most computational burden of adaptation via a feedback model is on the forward computation of either the pre-defined or pre-trained feedback model based on the task context and sensory inputs. Hence, reactivity via a feedback model is computationally cheaper \changemarker{online} than reactivity via a re-planning method \citep{Eppner_LessonAPC_RSS16, Kappler_RTPerceptionReactiveMotion_RSS17}.

Moreover, adaptation via re-planning requires the ability to plan ahead, which in turn requires a model that can predict the consequences of planned actions. For cases such as movement in free space and obstacle avoidance, such predictive model is straightforward to derive, and hence re-planning approaches are feasible in these cases --although possibly still computationally expensive. On the other hand, for tasks involving interaction with the environment, for instance when trying to control contacts through tactile sensing, planning is hard due to hard-to-model non-rigid contact dynamics as well as other non-linearities. In this case, learning methods may come to the rescue. Planning as well as re-planning approaches are possible once the predictive model is learned \citep{TianManipulationByFeelICRA19, hoffmann2014adaptive, icra2019_learning_tactile_servoing}. Similarly, a reactive policy via a \textit{learned} feedback model is also a possible solution to achieve adaptation in this case \citep{Chebotar_IROS_2014, icra2018_learn_tactile_feedback}. However, in order to plan and re-plan effectively, a \textit{global} predictive model \changemarker{--global in the sense that it covers the whole state space--} is required. On the other hand, feedback models are supposed to react to changes locally, thus \textit{local} feedback models are sufficient. \changemarker{Although \textit{local} models most of the time only achieve sub-optimality, often the solution is good enough to accomplish the task. Furthermore,} learning a \textit{local} feedback model is more sample-efficient as compared to learning a \textit{global} predictive model for a re-planning approach, and hence we choose to work on the former in this paper.

Among the data-driven approaches for learning reactive policies, there are at least two alternatives. One can learn a global end-to-end sensing-to-action policy. 
Alternatively, we can provide a nominal motion plan encoded as a movement primitive, and then learn a local feedback adaptation model on top of the primitive to account for environmental changes, along the same line of thought as the residual learning \citep{Johannink_Residual_RL_Robot_Control}.
Although the global end-to-end reactive policy learning has shown some impressive results, such as learning reactive policy mapping raw images to robot joint torques \citep{levine_e2e_deep_visuomotor_policies, Chebotar_PIGPS}, we chose to work on the latter, as 
the representation with movement primitives and local feedback models explicitly incorporate task parameterization such as start, goal, and movement duration, making this method more generalizable to variations of these parameters. Specifically, we utilize Dynamic Movement Primitives (DMP) \citep{Ijspeert_NC_2013} to represent the nominal motion. DMPs allow the possibility of adding reactive adaptations of the motion plan based on sensory input \citep{pastor_IROS_2011_ASM}.
This modulation is realized through the so-called \textit{coupling terms} which can be a function of the sensory input, essentially creating a feedback loop. This means the DMP becomes a reactive controller. Such online modulation is generated by a feedback model, which maps sensory information of the environment to the motion plan adaptation.

In the past, robotics researchers have been hand-designing feedback models for particular applications \citep{Park_Humanoids_2008, Hoffmann_ICRA_2009}. However, such feedback models are very problem-specific and most likely are not re-usable in another problem. Moreover, in cases where sensory input is very rich and high-dimensional, mapping this input into the required adaptation is not straightforward, and hand-designing the feedback model can become a tedious process. For this reason, we have proposed data-driven approaches to learning feedback models \citep{icra2017_learning_feedback, icra2018_learn_tactile_feedback}.
In this previous work, we acquired the feedback model in a supervised manner, by learning from demonstrations. 

This manuscript extends our prior work in several ways: First, we present an extended, more complete, version of our prior work on learning feedback models based on tactile sensing \citep{icra2018_learn_tactile_feedback}. This extended version includes details on our tactile feedback testbed; details and insights on how to acquire demonstrations for skills involving tactile sensing and a method for semi-automatic demonstration segmentation. Second, to allow for generalization to previously unseen task settings, we extend our work with a sample-efficient reinforcement learning (RL) approach. 
\begin{figure}[ht]
	\centering
    \includegraphics[width=1.0\columnwidth]{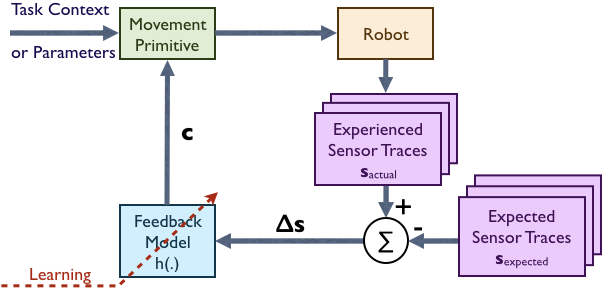}
    \caption{Proposed framework for learning feedback models with Associative Skill Memories for reactive motion planning.}
    \label{fig:BlockDiagramLearnFbTermsASM}
\end{figure}
Once the feedback model has been initialized by learning-from-demonstration on several known environmental settings, we tackle the problem of encountering new environment settings during execution. In such new settings, we propose an RL extension of the framework, allowing the feedback model to be refined further by trial-and-error such that the performance of the feedback model on the new setting can be improved over a few trials, while maintaining its performance on the settings it has been trained on before. The new setting can be selected from an extrapolated environment setting --instead of an interpolated one-- over the known settings. This gives the system the possibility for constantly expanding the possible range of environmental settings that it can handle, leading toward a lifelong learning of adaptation in reactive behaviors.

We perform our experimental evaluation on a learning task for tactile feedback on a real robot. 
In this tactile feedback testbed, we are learning the mapping from the sensor traces deviation $\sensortracesdeviation$ -- the difference between the experienced sensor traces $\sensoract$ and the expected one $\sensorexp$, according to the concept of Associative Skill Memories (ASMs) \citep{pastor_IROS_2011_ASM}-- to the adaptation $\couplingterm$ of the behavior. Data efficiency is critical in this case, given that the tactile-driven action dynamics is not easy to be simulated and the data collection on the real hardware is expensive. Thus this helps us to test how feasible is our framework of learning feedback model, especially when applied to the real robot scenario.
Our proposed framework is depicted in Figure \ref{fig:BlockDiagramLearnFbTermsASM}.

\section{Related Work}
\label{sec:rlfb_related_work}
There are several pieces of related work \citep{Kappler_RTPerceptionReactiveMotion_RSS17, cheng2018rmpflow, RatliffRMP} that \changemarker{have} tried to combine re-planning \changemarker{with} reactive feedback models, in order to get the best out of both approaches: \changemarker{the ability} to instantly adapt to local changes, \changemarker{while also being able} to re-plan whenever more significant changes occur. In this paper, we focus our work on the general learning of representations for adaptation via feedback models which has not been explored significantly in the past. We start out by reviewing motion representations.
\subsection{Movement Representation}
\label{ssec:rlfb_related_work_movement_representation}
Some previous projects in the literature encoded movement plans as primitives for modular motion planning \citep{Ijspeert_NC_2013, Paraschos_ProMP, Ewerton_ProMP_DiffSpeedsExec}. Dynamical Movement Primitive (DMP) with local coordinate transformation \citep{Ijspeert_NC_2013} and Associative Skill Memories (ASMs) \citep{pastor_IROS_2011_ASM, pastor2013dynamic} pose a framework for data-driven modular sensory-motor primitives. It is modular in the sense that the learned primitive can generalize w.r.t. task parameters such as start, goal, and movement duration. 

The parameters of motion primitives can be learned via learning from demonstrations or via RL. 
One of the advantages of DMPs is in the ability to represent a continuous motion trajectory with a small number of parameters, typically of size 25-50 parameters per task space dimension. For such a small parameterization, gradient-free policy-based RL methods such as PoWER \citep{Kober_PoWER} or PI\textsuperscript{2} \citep{Theodorou_PI2} are sample-efficient, and have been successfully used in many applications \citep{Stulp_2012_DMP_RL, Kalakrishnan_LearnForceControlPolicies, Hazara_PI2_DMP_InContactSkills, Celemin2019IJRR_RL_MP_PolicySearch_CorrectiveAdvice}, to learn or adjust the nominal behaviors within a few RL iterations.
\subsection{Automated Segmentation of Demonstrations}
For learning from human demonstrations, after demonstrations are provided, a pre-processing step is required to segment each demonstration into the corresponding primitive. This segmentation is based on the function/role and semantics of each primitive in the task, i.e. as parts of a movement primitive library which can be re-used for other complex tasks such as writing or assembly tasks \citep{Meier_movement_segmentation_library_IROS11, Niekum_SemanticSegmentation_RSS13, niekum2015learning, lioutikov2015probabilistic}.

The simplest technique for demonstration segmentation is the threshold-based cutting, for example based on a threshold on the velocity norm signal like in the Zero Velocity Crossing (ZVC) method \citep{fod2002automated}. While this method 
--in combination with manual inspection of the segmentation result-- 
is sufficient for segmenting a small number of demonstrations, for a larger number of demonstrations this method becomes tedious since we need to find a threshold that will work for all noisy demonstrations.

A more sophisticated segmentation method is based on feature correspondence matching between demonstrations, which is done by dynamic programming techniques such as Dynamic Time Warping (DTW) \citep{Sakoe_ASSP_1978, Coates_LearningInvertedHelicopter_ICML08, Zhou_GTW_CVPR12, Aung_TrajAlign_ICPR10}. However, due to the noise in the demonstrations, the extracted feature correspondences can be erroneous, which in turn will degrade the segmentation quality.

In this paper, we introduce a more robust method for automated demonstration alignments and segmentation by combining the feature correspondences extracted by DTW with a weighted least square formulation of the problem, in a similar nature as the Iterative Closest Point (ICP) method \citep{Chen_IVC_1992} in Computer Graphics for mesh alignment and registration.
\subsection{Learning of Feedback Models}
Previously many approaches have hand-designed feedback models for specific purposes, such as obstacle avoidance \citep{Park_Humanoids_2008, Hoffmann_ICRA_2009, Pastor_ICRA_2009, Zadeh_AutonRob_2012}, joint limit avoidance \citep{gams2009line}, as well as force and tactile feedback \citep{Khansari_IJRR_2016, pastor_IROS_2011_ASM, hogan2020tactiledexterity, donlon2018gelslim, BalaLearningTactileForceEstimationICRA19, VeigaGripStabilizationTactileFeedbackSensors20}.

However, the cumbersome parameter tuning becomes a drawback of hand-crafted feedback modeling approaches, which motivates more data-driven approaches. \cite{Johannink_Residual_RL_Robot_Control, Sung_ICRA_2017} learn feedback models for haptics and contact interaction with the environment. Although the methods are data-driven, there is no performance guarantees. Hence, the next question is about how to incorporate learning in the feedback model acquisition, while ensuring desirable system properties. 
\cite{rai2014learning, icra2017_learning_feedback, Pairet_IROS19} learn feedback models for obstacle avoidance with goal-convergence guarantees, either by designing features with Lyapunov stability proofs or by adding post-processing heuristics. 

Besides the convergence properties, previous works showed that the feedback models may also have dependencies on movement phases \citep{Kober_IROS_2008, Chebotar_IROS_2014}. 
Moreover, \cite{gams2015learning} employed Learning-from-Demonstration (LfD) to train separate feedback models for different environmental settings, which were then interpolated by a Gaussian Process regression. 

In our previous work \citep{icra2018_learn_tactile_feedback} and in this paper, we present a unified and general-purpose feedback model learning representation which has goal-convergence properties, can capture phase-dependent movement adaptation, and can handle multiple environmental settings.

\subsubsection{Reinforcement Learning of Feedback Models}
~\\
Previous works \citep{Kober_IROS_2008, Chebotar_IROS_2014} are mostly performing RL to acquire feedback policies with low number of parameters or are limited to linear weighted combination of phase-modulated features.

Deep RL techniques \citep{Mnih_DQN, Lillicrap_DDPG, Tamar_VIN} offer a solution for learning policies with high number of parameters, such as those represented by neural networks. 
Although Deep RL techniques have shown promising results in simulated environments, the high sample complexity of these methods hinder their use in real-world robot learning involving physical hardware. When simulations are available, Simulation-to-Real methods \citep{Chebotar_Sim2Real, Molchanov_Sim2Real} can be leveraged to perform high sample complexity training in simulation, while only updating the dynamics model in the simulation from real-world rollouts a few times. However, for our tactile feedback testbed in this paper, the dynamics models are difficult to be obtained or simulated, and hence we really need a sample-efficient RL for our purpose.

Guided Policy Search (GPS) \citep{levine_gps} tackles the high sample complexity issue by decomposing the policy search into two parts: trajectory optimization and supervised learning of the high-dimensional policy. However, the use of smooth local policies in GPS --such as LQR with local time-varying linear models-- to supervise the high-dimensional policy has difficulties in learning discontinuous dynamics, such as during a door opening task. Path Integral Guided Policy Search (PIGPS) \citep{Chebotar_PIGPS} replaced LQR with PI\textsuperscript{2} --a model-free and gradient-free RL algorithm-- to tackle the problem of learning discontinuous dynamics.

\section{Review: Dynamical Movement Primitives}
\label{sec:rlfb_background_dmp}
Dynamical Movement Primitives (DMPs) \citep{Ijspeert_NC_2013} is a goal-directed behavior described as a set of differential equations with well-defined attractor dynamics.
Moreover, for the purpose of enabling online adaptation via feedback models, DMPs allow modulation of sensory inputs as a feedback adaptation to the output motion plans, via an additional term in the differential equations.

In our framework, we use DMPs to represent position and orientation motion plans of the robot's end-effector, as well as to represent the expected sensory traces in the Associative Skill Memories (ASM) framework \citep{pastor2013dynamic}. As mentioned in Section \ref{sec:rlfb_intro}, the expected sensory traces $\sensorexp$ are the reference signals of the adaptation feedback model. Throughout this manuscript, we call the DMP models that we use to represent position motion plans and expected sensory traces as \textit{regular} DMPs. On the other hand, we use quaternions to represent orientations, and hence we call the orientation DMP models as \textit{Quaternion} DMPs. 
\cite{icra2017_learning_feedback} provides a reference for our regular DMP formulation, while here we focus on reviewing Quaternion DMPs.
Quaternion DMPs were first introduced in \citep{pastor_IROS_2011_ASM}, and then improved in \citep{Ude_ICRA14_OrientationDMP, Kramberger_Humanoids16_GenOrientationDMP} to fully take into account the geometry of the Special Orthogonal Group $\textit{SO}$(3).

In general, a DMP model consists of a \emph{transformation system}, a \emph{canonical system}, and a \emph{goal evolution system}. The \textit{transformation system} governs the evolution of the state variable being planned, which for a Quaternion DMP is defined as\footnote{For defining Quaternion DMPs, the operators $\circ$, ${}^{*}$, the logarithm mapping $\log(\cdot)$, and the exponential mapping $\exp(\cdot)$ are required. The definition of these operators are stated in Equations~\ref{eq:QuatComposition}, \ref{eq:QuatConjugation}, \ref{eq:LogMapping}, and \ref{eq:ExpMapping} in the Appendix \ref{ap:quaternion_algebra}.}:
\begin{equation}
    \dmpmotiondurationprop^2 \angularacceleration = \alpha_{\omega} \left( \beta_{\omega} 2 \log\left( \quatevolvinggoalstateposition \circ \quatstateposition^{*} \right) - \dmpmotiondurationprop \angularvelocity\right) + \quatforcingterm + \quatcouplingterm
    \label{eq:OriDMPTransformationSystem}
\end{equation}
where $\quatstateposition$ is a unit quaternion representing the orientation, $\quatstateposition_g$ is the goal orientation, and $\angularvelocity, \angularacceleration$ are the 3D angular velocity and angular acceleration, respectively. $\quatforcingterm$ and $\quatcouplingterm$ are the 3D orientation \textit{forcing term} and \textit{feedback/coupling term}\footnote{Throughout this manuscript, we use the term \emph{feedback} and the term \emph{coupling term} interchangeably.}, respectively.

During unrolling, we integrate $\quatstateposition$ forward in time to generate the kinematic orientation trajectory as follows:
\begin{equation}
    \quatstateposition_{t+1} = \exp\left( \frac{\angularvelocity \Delta t}{2} \right) \circ \quatstateposition_{t}
    \label{eq:QuatIntegration}
\end{equation}
where $\Delta t$ is the integration step size.
The forcing term encodes the nominal behavior, while the coupling term encodes behavior adaptation which is commonly based on sensory feedback. Our work focuses on learning feedback models that generate the coupling terms.

We choose the constants 
$\alpha_{\omega} = 25$ and 
$\beta_{\omega} = \alpha_{\omega}/4$ 
to achieve a critically-damped system response when both forcing term and coupling term are zero \citep{Ijspeert_NC_2013}. $\dmpmotiondurationprop$ is set proportional to the motion duration.

The second-order \textit{canonical system} governs the movement phase variable $\phasevariable$ and phase velocity $\phasevelocity$ as follows:
\begin{equation}
	\dmpmotiondurationprop \dot{\phasevelocity} = \alpha_{\phasevelocity} \left( \beta_{\phasevelocity} \left( 0 - \phasevariable \right) - \phasevelocity\right)
	\label{eq:2ndOrderCanonicalSystemP1}
\end{equation}
\begin{equation}
	\dmpmotiondurationprop \dot{\phasevariable} = \phasevelocity
	\label{eq:2ndOrderCanonicalSystemP2}
\end{equation}
We set the constants $\alpha_{\phasevelocity} = 25$ and $\beta_{\phasevelocity} = \alpha_{\phasevelocity}/4$ to get a critically-damped system response. The phase variable $\phasevariable$ is initialized with 1 and will converge to 0. On the other hand, the phase velocity $\phasevelocity$ has initial value 0 and will converge to 0. For the purpose of our work, position DMPs, Quaternion DMPs and expected sensory traces share the same canonical system, such that the transformation systems of these DMPs are synchronized \citep{Ijspeert_NC_2013}.

The \textit{forcing term} $\forcingterm$ governs the shape of the transient behavior of the primitive towards the goal attractor. The forcing term is represented as a weighted combination of $N$ basis functions $\psi_i$ with width parameter $\chi_i$ and centered at $\mu_i$:
\begin{equation} 
  \forcingterm
  \left( \phasevariable, \phasevelocity; \dmpparamset \right) = \frac{\sum_{i=1}^N \psi_i \left( \phasevariable \right) \dmpparam_i}{\sum_{j=1}^N \psi_j \left( \phasevariable \right)} \phasevelocity
  \label{eq:DMPForcingTerm}
\end{equation}
where
\begin{equation} 
  \psi_i \left( \phasevariable \right) =  \exp\left( -\chi_i \left( \phasevariable - \mu_i \right)^2 \right)
  \label{eq:GaussianBasisFunction}
\end{equation}
Note, because the forcing term 
$\forcingterm$ 
is modulated by the phase velocity $\phasevelocity$, it is initially $0$ and will converge back to $0$.

The $N$ basis function weights $\dmpparam_i$ in equation \ref{eq:DMPForcingTerm} are learned from human demonstrations of \textit{baseline/nominal behaviors}, by setting the target regression variable:
\begin{align*}
    \targetquatforcingterm = &-\alpha_{\omega} (\beta_{\omega} 2 \log\left( {\quatevolvinggoalstateposition}_{, \text{bd}} \circ \quatstateposition_\text{bd}^{*} \right) - \dmpmotiondurationprop_\text{bd} \angularvelocity_\text{bd}) \\
    &+ \dmpmotiondurationprop_\text{bd}^2 \angularacceleration_\text{bd}
    \numberthis
    \label{eq:OriDMPForcingTermExtraction}
\end{align*}
where \{$\quatstateposition_\text{bd}, \angularvelocity_\text{bd}, \angularacceleration_\text{bd}$\} is the set of \textit{baseline/nominal} orientation behavior demonstrations.
$\dmpmotiondurationprop_\text{bd}$ is the movement duration of the baseline/nominal behavior demonstration.
From this point, we can perform linear regression to identify parameters $\dmpparamset$, as shown in \citep{Ijspeert_NC_2013}.

Finally, we include a \textit{goal evolution system} as follows:
\begin{equation}
    \dmpmotiondurationprop \angularvelocity_g = \alpha_{\omega_{g}} 2 \log \left( \quatsteadygoalstateposition \circ \quatevolvinggoalstateposition^{*} \right) 
    \label{eq:OriDMPGoalEvolSystem}
\end{equation}
where $\quatevolvinggoalstateposition$ and $\quatsteadygoalstateposition$ are the evolving and the steady-state goal orientation, respectively.
We set the constant 
$\alpha_{\omega_{g}} = \alpha_{\omega}/2$.
The goal evolution system has two important roles related to safety during the algorithm deployment on robot hardware. The first role, as mentioned in \citep{Ijspeert_NC_2013}, is to avoid discontinuous jumps in accelerations when the goal is suddenly moved. 
The second role, as mentioned in \citep{nemec2012action}, is to ensure continuity between the state at the end of one primitive and the state at the start of the next one when executing a sequence of primitives. 
Here we ensure continuity between primitives for both position and orientation DMPs by adopting \citep{nemec2012action}.

In the upcoming experimental evaluation on the learning tactile feedback testbed, we will be using a sequence of three primitives, therefore we first explain our method for segmenting the nominal behavior demonstrations into the corresponding primitives. 
\section{Overview: Learning Feedback Models for Reactive Behaviors}
\label{sec:lfb_overview}
\begin{figure}[ht]
	\centering
    \includegraphics[width=0.8\columnwidth]{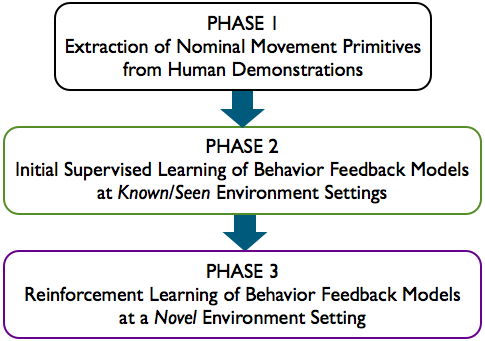}
    \caption{Flow diagram of our framework. Phase 1 is detailed in section~\ref{sec:lfb_extraction_of_movement_primitives_w_auto_demo_segmentation} as outlined in Algorithm~\ref{alg:extraction_nominal_mp_from_demos}. Phase 2 is detailed in section~\ref{sec:lfb_model} with its flow diagram expanded in Figure~\ref{fig:FlowDiagramLearnFbTerms_Supervised}. Phase 3 is detailed in section~\ref{sec:lfb_rl} as outlined in Algorithm~\ref{alg:RLFBreactiveMP}.}
	\label{fig:FlowDiagramLearnFbTerms_Overall}
\end{figure}
We aim to realize reactive behavior adaptations via feedback models \changemarker{given 
sensory observations} within one unified machine learning framework.
We envision a general-purpose feedback model learning representation framework that can be used in a variety of scenarios, such as for obstacle avoidance, tactile feedback, locomotion planning adjustment, etc. Such a learning representation ideally has the flexibility to incorporate a variety of sensory feedback signals and can be initialized from human demonstrations. Finally, our learning framework should be able to perform reinforcement learning to refine the model when encountering new environmental settings, while maintaining the performance on the previously seen settings.

Our overall learning framework contains 3 main phases: In \textsc{Phase 1}, we start out by collecting human demonstrations of default/\emph{nominal} behaviors for a task executed in a \emph{default} environmental setting. These demonstrations represent the motion trajectories and sensory traces of successful executions. We semi-automatically segment the demonstrations into several movement primitives and encode both the default motion trajectory as well as the \emph{expected sensor traces} of a given segment into DMPs. We describe this phase in detail in section~\ref{sec:lfb_extraction_of_movement_primitives_w_auto_demo_segmentation}. In \textsc{Phase 2} we then unroll the learned movement primitives on a variety of non-default environmental settings. 
Due to the change in the environment setting, it is necessary to adapt/correct the behaviors in order to accomplish the task. Thus we provide demonstrations of how to correct each primitive execution. Simultaneously, we also record the \emph{experienced} sensor traces of these corrected rollouts, followed by a supervised learning of behavior feedback models from these correction demonstrations. We describe the details of this phase in section~\ref{sec:lfb_model}. In \textsc{Phase 3} we now turn to learning how to correct without demonstrations in novel environmental settings. We propose a sample-efficient reinforcement learning algorithm, that can update the feedback model such that behavior is improved on novel settings as well as maintaining its performance on previously known/seen settings. We present the details of our reinforcement learning phase in section~\ref{sec:lfb_rl}. \changemarker{
While it is possible to skip \textsc{Phase 2}, it has been shown that bootstrapping the training by learning from human demonstrations can increase sample-efficiency of the RL phase \citep{levine_gps, Chebotar_PIGPS, Sung_ICRA_2017}.} An overview of our framework, with its phases, is shown in Figure~\ref{fig:FlowDiagramLearnFbTerms_Overall}.

\subsection{Testbed: Tactile Feedback}
\begin{figure}[t]
  \centering
  \includegraphics[trim={0.0cm 1.25cm 0.0cm 6.0cm},clip,width=0.6\columnwidth]{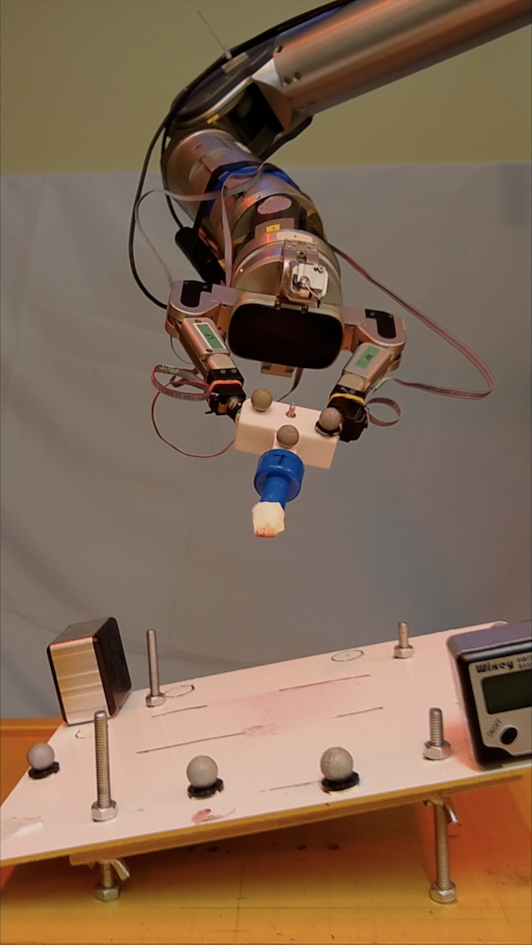}
  \caption{Learning tactile feedback testbed.}
  \label{fig:rlfb_learning_tactile_feedback_setup}
\end{figure}

We perform evaluations of our framework on a learning tactile feedback testbed as depicted in Figure~\ref{fig:rlfb_learning_tactile_feedback_setup}. In this testbed, the robot is equipped with two tactile sensors at its fingers. These tactile fingers perform pinch grasp on a scraping tool, and the task is to scrape the board flatly w.r.t. to the board's surface. The challenge is that the robot does not know a-priori what is the orientation of the board --which is a representation of the environment setting--, since it is not given a visual information of the board. The robot needs to online correct the execution of the scraping task --by relying on the tactile sensing information-- in order to still successfully execute the task. In this testbed, we are acquiring/learning the feedback model which maps the tactile sensing information $\sensortracesdeviation$ --as a proxy of the environment setting information-- to the correction/adaptation $\couplingterm$ of the behavior.
The division of the task into movement phases that are considered as movement primitives is usually dependent on the function/role and semantics of the movement phase with respect to the overall task. In the learning tactile feedback testbed that we are evaluating in this paper and in our previous work \citep{icra2018_learn_tactile_feedback}, the task is broken down into three primitives: bringing the tool down until contact is made with the scraping board, tool re-orientation to align the tool tip with the board surface, and scraping forward on the board.

\section{Segmenting Demonstrations into Movement Primitives}
\label{sec:lfb_extraction_of_movement_primitives_w_auto_demo_segmentation}
In our work, we model complex manipulation tasks as a sequence of motion primitives. Moreover, for robustness we want to learn these primitives from multiple demonstrations of the \textit{baseline/nominal} behavior. In a learning-from-demonstrations setup, it is often easier to obtain each demonstration for such tasks in one go, instead of one primitive at a time. Thus, our work starts by collecting such full demonstrations, and then segmenting the recordings into motion primitives. 

\changemarker{To avoid the high cost of manually segmenting all demonstrations, we propose a semi-automated segmentation approach. The main idea of our semi-automated approach is that a user picks one demonstration and manually segments it, and then our approach segments the rest of the demonstrations by aligning it with the manually segmented one. While not fully automatic, only having to manually segment one demonstration is better than having to manually segment all demonstrations.} 

\changemarker{Our approach starts with someone  manually segmenting a} demonstration $\firstunsegmentedtraj = \{\unsegmentedtrajdatapoint_i\}_{i=1}^{T_1}$ in our dataset into $\numprimitives$ trajectories $\{\unsegmentedtrajdatapoint_i\}_{i=\primretrajstartidx}^{\primretrajendidx}$, one per primitive $p$. For this manual segmentation we use the Zero Velocity Crossing (ZVC) method \citep{fod2002automated}, with a manually tuned threshold $\zvcthresh$.
Moreover, we expand each segmented trajectory in both directions -- the start and end points -- by $\dtwsearchspaceextension$, and call the result the \textit{reference} trajectory $\retraj^\idxprimitive = \{\unsegmentedtrajdatapoint_i\}_{i=\primretrajstartidx-\dtwsearchspaceextension}^{\primretrajendidx+\dtwsearchspaceextension}$. 

Given, these $\numprimitives$ reference trajectories, the automatic segmentation of the unsegmented demonstrations is outlined in Algorithm~\ref{alg:extraction_nominal_mp_from_demos}. For a task that consists of $\numprimitives$ primitives, we incrementally segment one primitive at a time. The segmentation process takes the following as input:
\begin{itemize}
    \item the remaining $(\numbaselinedemos - 1)$ un-segmented \textit{baseline} demonstrations $\{\indexedunsegmentedtraj\}_{\idxbaselinedemo=2}^{\numbaselinedemos}$
    \item the segmented reference trajectories $\{\retraj^\idxprimitive\}_{\idxprimitive=1}^{P}$
    \item Zero Velocity Crossing (\zvcfunction) threshold $\zvcthresh$.
    \changemarker{We use the same velocity threshold $\zvcthresh$ as the one used for segmenting the reference trajectory $\retraj^\idxprimitive$.}
    \item Dynamic Time Warping (\dtwfunction) search's integer time index extension $\dtwsearchspaceextension$
\end{itemize}
\begin{algorithm}[ht]
	\caption{Pseudo-Code of Nominal Movement Primitives Extraction from Demonstrations}
	\label{alg:extraction_nominal_mp_from_demos}
	\begin{algorithmic}[1]
	    \Function{SegmentDemo}{$\{\indexedunsegmentedtraj\}_{\idxbaselinedemo=2}^{\numbaselinedemos}$, $\{\retraj^\idxprimitive\}_{\idxprimitive=1}^{P}$, $\zvcthresh$, $\dtwsearchspaceextension$}
    		\State \# Initialize solution with empty list for $\numprimitives$ primitives:
    		\State $\extractednominalprimslist = \{\{\} \times \numprimitives\}$
            \For{$\idxprimitive = 1$ to $\numprimitives$}
                \For{$\idxbaselinedemo = 2$ to $\numbaselinedemos$}
                    \State \# Compute the \textit{initial segmentation guess}
                    \State \# start point $\primtargettrajstartidx$ and end point $\primtargettrajendidx$ with ZVC:
                    \State $[\primtargettrajstartidx, \primtargettrajendidx] = \zvcfunction(\indexedunsegmentedtraj, \zvcthresh)$ \label{aln:targettraj_initial_segmentation_guess_w_zvc}
                    \State {}
                    \State \# If $\indexedunsegmentedtraj = \{\unsegmentedtrajdatapoint_j\}_{j=1}^{T_\idxbaselinedemo}$, then:
                    \State ${\targettraj}^\idxprimitive = \{\unsegmentedtrajdatapoint_j\}_{j=\primtargettrajstartidx-\dtwsearchspaceextension}^{\primtargettrajendidx+\dtwsearchspaceextension}$ \label{aln:targettraj_definition}
                    \State {}
                    \State \# Compute correspondence pairs $\correspondencepairslist$:
                    \State $\correspondencepairslist = \dtwfunction({\retraj}^\idxprimitive, {\targettraj}^\idxprimitive)$ \label{aln:dtw_correspondence_extraction}
                    \State {}
                    \State \# Construct least square (LS) parameters:
                    \State $[\mathbf{A}, \mathbf{b}] = \lsconstructfunction(\correspondencepairslist)$ \label{aln:ls_construction}
                    \State {}
                    \State \# Compute least square (LS) weights:
                    \State $\mathbf{W}= \lswcomputefunction(\correspondencepairslist, {\retraj}^\idxprimitive, {\targettraj}^\idxprimitive)$ \label{aln:ls_weight_heuristics_computation}
                    \State {}
                    \State \# Solve weighted least square (WLS) problem
                    \State \# to estimate time scale $\targettimescale$ and time delay $\targettimedelay$:
                    \State $[\targettimescale, \targettimedelay] = \wlssolvefunction(\mathbf{A}, \mathbf{b}, \mathbf{W})$ \label{aln:wls_solution}
                    \State {}
                    \State \# Append the refined segmentation to solution:
                    \State $i_s = \primtargettrajstartidx+\targettimedelay+(\targettimescale-1)\dtwsearchspaceextension$
                    \State $i_e = \primtargettrajstartidx+\targettimedelay+(\targettimescale-1)\dtwsearchspaceextension+\targettimescale(\primretrajendidx-\primretrajstartidx)$
                    \State $\extractednominalprimslist[\idxprimitive] = \extractednominalprimslist[\idxprimitive] \cup \{\{\unsegmentedtrajdatapoint_j\}_{j=i_s}^{i_e}\}$ \label{aln:targettraj_refinement}
                \EndFor
            \EndFor
            \State \Return $\extractednominalprimslist$
        \EndFunction
	\end{algorithmic}
\end{algorithm}

Our algorithm starts by extracting \textit{initial segmentation guesses} ${\targettraj}^\idxprimitive$ using the ZVC method from the remaining demonstrations. However, we expect these initial segments to be very rough, and thus expand each initial segmentation guess to $\dtwsearchspaceextension$ time steps before and after the found start and end points\footnote{In practice, when extending the trajectories --e.g. as done in line~\ref{aln:targettraj_definition} of Alg.~\ref{alg:extraction_nominal_mp_from_demos}--, we need to take care of boundary cases, because the extended segmentation points may fall out-of-range. However, for the sake of clarity, here we do not include this boundary-checking in Algorithm~\ref{alg:extraction_nominal_mp_from_demos}.}.

For each primitive $p$, we then use Dynamic Time Warping (DTW) \citep{Sakoe_ASSP_1978} to extract point correspondences $\correspondencepairslist$ between the reference trajectory ${\retraj}^\idxprimitive$ and the guessed trajectory segment ${\targettraj}^\idxprimitive$, as shown in line~\ref{aln:dtw_correspondence_extraction} of Alg.~\ref{alg:extraction_nominal_mp_from_demos} and explained in section~\ref{ssec:lfb_auto_demo_segmentation_point_correspondence_matching_via_dtw}. 
Next, using the correspondence pairs $\correspondencepairslist$, we want to estimate a time delay and a time scaling parameter that best aligns the correspondence points in ${\targettraj}^\idxprimitive$ with its correspondence pair in ${\retraj}^\idxprimitive$, as we explain in section~\ref{ssec:lfb_auto_demo_segmentation_ls_construction}.
We propose to use a (weighted) least squares formulation to identify these parameters, which then allow us to extract the final segmentation, as will be explained in section~\ref{ssec:wls_problem_setup}. The overall segmentation algorithm is shown in Alg.~\ref{alg:extraction_nominal_mp_from_demos}. Next, we describe the details of the individual segmentation steps.

\subsection{Point Correspondences between Demonstrations}
\label{ssec:lfb_auto_demo_segmentation_point_correspondence_matching_via_dtw}
For finding point correspondences between the reference and guessed segmentation trajectories, we use the Dynamic Time Warping (DTW) algorithm \citep{Sakoe_ASSP_1978}\footnote{Please note that it is important to standardize both the reference and target trajectories before performing the DTW matching, such that each signal is zero-mean and its standard deviation equal to one, otherwise the computed correspondence pairs will be erroneous due to the incorrect/biased current cost.}. 

With reference to line~\ref{aln:targettraj_definition} of Algorithm~\ref{alg:extraction_nominal_mp_from_demos}, given:
\begin{enumerate}
    \item the one-dimensional (1D) reference trajectory ${\retraj}^\idxprimitive = \{\unsegmentedtrajdatapoint_i\}_{i=\primretrajstartidx-\dtwsearchspaceextension}^{\primretrajendidx+\dtwsearchspaceextension}$ of size $N = \primretrajendidx - \primretrajstartidx + 2\dtwsearchspaceextension + 1$
    \item a 1D initial segmentation guess trajectory ${\targettraj}^\idxprimitive = \{\unsegmentedtrajdatapoint_j\}_{j=\primtargettrajstartidx-\dtwsearchspaceextension}^{\primtargettrajendidx+\dtwsearchspaceextension}$ of size $M = \primtargettrajendidx - \primtargettrajstartidx + 2\dtwsearchspaceextension + 1$
\end{enumerate}
the DTW algorithm returns $K = \text{minimum}(M,N)$ correspondence pairs $\correspondencepairslist = \{(\correspondencepointre_k, \correspondencepointtarget_k)\}_{k=1}^{K}$, where $1 \leq \correspondencepointre_k \leq N$ is a (positive integer) time index in the reference trajectory and $1 \leq \correspondencepointtarget_k \leq M$ is a (positive integer) time index in the guessed segment.

\subsection{Least Square Problem Setup}
\label{ssec:lfb_auto_demo_segmentation_ls_construction}
Similar to the Iterative Closest Point (ICP) method \citep{Chen_IVC_1992} in Computer Graphics for mesh alignment, we can setup a least square problem for trajectory segment alignment. To do so, we relate each correspondence pair in $\correspondencepairslist = \{(\correspondencepointre_k, \correspondencepointtarget_k)\}_{k=1}^{K}$ by assuming that the time indices $\correspondencepointre_k$ in the reference segment can be shifted and scaled such that they match with the corresponding time indices $\correspondencepointtarget_k$ in the guessed segment:
\begin{equation}
	\correspondencepointtarget_k = \targettimescale \correspondencepointre_k + \targettimedelay
	\label{eq:tartraj_model}
\end{equation}
We would like to do parameter identification for $\targettimescale$ and $\targettimedelay$, which are the time scale and the time delay, respectively, of the guessed segment $\targettraj$ relative to the reference segment $\retraj$. 
Since the reference trajectory $\retraj$ has been properly cut using the ZVC method, knowing the values of $\targettimescale$ and $\targettimedelay$ parameters informs the best (data-driven) refined segmentation point of the guessed trajectory $\targettraj$.
To identify parameters $(\targettimescale, \targettimedelay)$ we can set up a least square estimate by using all $K$ correspondence pairs as follows:
\begin{equation}
	\mathbf{A} \mathbf{x} = \mathbf{b}
\end{equation}
with:
\begin{flalign*}
	\mathbf{A} =
	\begin{bmatrix}
		\correspondencepointre_1 & 1\\
		\correspondencepointre_2 & 1\\
		\vdots & \vdots\\
		\correspondencepointre_K & 1
	\end{bmatrix},
	\mathbf{b} =
	\begin{bmatrix}
		\correspondencepointtarget_1\\
		\correspondencepointtarget_2\\
		\vdots\\
		\correspondencepointtarget_K
	\end{bmatrix},
	\mathbf{x} =
	\begin{bmatrix}
		\targettimescale\\
		\targettimedelay
	\end{bmatrix}
\end{flalign*}
With $\mathbf{A}$ and $\mathbf{b}$ known, the regular solution of least square applies:
\begin{equation}
	\mathbf{x} = \left(\mathbf{A}\T\mathbf{A}\right)^{-1}\mathbf{A}\T\mathbf{b}
	\label{eq:LS_solution}
\end{equation}

\subsection{Weighted Least Square Solution}
\label{ssec:wls_problem_setup}
~\\
The least square solution in Equation \ref{eq:LS_solution} above applies if the assumption --that all correspondence pairs are accurate-- is satisfied. In reality, due to noise, un-modeled disturbances and other factors, some correspondence pairs can be inaccurate, yielding an inaccurate estimation of $\targettimescale$ and $\targettimedelay$ parameters. These inaccurate correspondence pairs can be characterized as follows:

\begin{itemize}
	\item correspondence pair between points with near-zero velocities
	\item correspondence pair between points of incompatible velocities
\end{itemize}

Correspondence pairs with near-zero velocities are most likely inaccurate, due to the way the DTW algorithm works. To give an intuitive example, let us imagine we have $\retraj$ which has a consecutive of 100 zero data points in the end (i.e. trailing zeroes), while $\targettraj$ has only 10 zero data points in the end. In this case, there are many number of ways the matching can be done between the 100 trailing zeroes in $\retraj$ with the 10 trailing zeroes in $\targettraj$. However, there is only one matching that will be consistent with the true value of the $(\targettimescale, \targettimedelay)$ parameters. 
Therefore, we filter out the correspondence pairs with near-zero velocities, so that they do not affect the $(\targettimescale, \targettimedelay)$ parameter identification result at all.

For handling the case of correspondence pairs between points of incompatible velocities, instead of using the regular least square solution, we use a weighted least square formulation as follows:
\begin{equation}
	\mathbf{x} = \left(\mathbf{A}\T\mathbf{W}\mathbf{A}\right)^{-1}\mathbf{A}\T\mathbf{W}\mathbf{b}
	\label{eq:WLS_solution}
\end{equation}
The weight matrix $\mathbf{W}$ is chosen to be a diagonal positive definite matrix, with each diagonal component is the weight associated with a correspondence pair. This weight is determined by how compatible the velocities are between the two points. 
If the velocities of the pair are the same, the weight will be 1 --which is the maximum weight possible--; otherwise, the weight decays exponentially as a function of the velocity difference.

We also performed time complexity analysis of our algorithm, which is $\mathcal{O}(MN)$ in overall, with $N$ is the length of the reference segment and $M$ is the length of the guessed segment as defined in section~\ref{ssec:lfb_auto_demo_segmentation_point_correspondence_matching_via_dtw}. A quadratic speed-up can be obtained by down-sampling the reference and guessed segments. Details of this time complexity analysis and speed-up can be seen in Appendix \ref{ap:auto_demo_segmentation_time_complexity_analysis_and_quadratic_speed_up}.

\subsubsection{Refinement of the Trajectory Segmentation}
\label{ssec:targettraj_segmentation_refinement}
~\\
After the time scale $\targettimescale$ and time delay $\targettimedelay$ parameters are identified, we can refine the segmentation of the guessed trajectory. The duration of the reference trajectory is $(\primretrajendidx-\primretrajstartidx+2\dtwsearchspaceextension)$. Hence, the segment in the guessed trajectory that corresponds to the reference trajectory has duration $\targettimescale(\primretrajendidx-\primretrajstartidx+2\dtwsearchspaceextension)$. Moreover, such segment starts at $\primtargettrajstartidx-\dtwsearchspaceextension+\targettimedelay$. Taking into account the scaling of the extension $\dtwsearchspaceextension$ as well, the refined trajectory segment that is to be appended to the solution has the start index:
\begin{equation}
    i_s = \primtargettrajstartidx-\dtwsearchspaceextension+\targettimedelay+\targettimescale\dtwsearchspaceextension = \primtargettrajstartidx+\targettimedelay+(\targettimescale-1)\dtwsearchspaceextension
\end{equation}
and the end index:
\begin{equation}
    \begin{aligned}
        i_e &= \primtargettrajstartidx-\dtwsearchspaceextension+\targettimedelay+\targettimescale(\primretrajendidx-\primretrajstartidx+2\dtwsearchspaceextension)-\targettimescale\dtwsearchspaceextension \\
        &= \primtargettrajstartidx+\targettimedelay+(\targettimescale-1)\dtwsearchspaceextension+\targettimescale(\primretrajendidx-\primretrajstartidx)
    \end{aligned}
\end{equation}
Eventually, we append the refined segmentation result to the solution, as shown in line~\ref{aln:targettraj_refinement} of Alg.~\ref{alg:extraction_nominal_mp_from_demos}.

\subsection{Learning DMPs from Multiple Segmented Demonstrations}
After we segmented the demonstrations into primitives, we can learn the forcing term parameters $\dmpparamset$ for each primitive from these segmented demonstrations. For each data points in each segmented demonstration, we extract the target regression variable $\targetforcingterm$ and the corresponding phase-dependent radial basis function (RBF) feature vector 
$\frac{\phasevelocity}{\sum_{j=1}^N \psi_j \left( \phasevariable \right)} 
\begin{bmatrix}
    \psi_1(\phasevariable) & \psi_2(\phasevariable) & \dots & \psi_N(\phasevariable)
\end{bmatrix}$ according to Eq.~\ref{eq:OriDMPForcingTermExtraction} and \ref{eq:GaussianBasisFunction}. Afterwards we stack these target regression variable $\targetforcingterm$ and the corresponding phase-dependent RBF feature vector over all the data points in all segmented demonstrations, and perform regression to estimate the forcing term parameters $\dmpparamset$ based on the relationship in Eq. \ref{eq:DMPForcingTerm}.

\section{Learning Feedback Models for Reactive Behaviors from Demonstrations}
\label{sec:lfb_model}
Recall the transformation system from Eq. \ref{eq:OriDMPTransformationSystem}:
\begin{align*}
    \dmpmotiondurationprop^2 \angularacceleration = \alpha_{\omega} \left( \beta_{\omega} 2 \log\left( \quatevolvinggoalstateposition \circ \quatstateposition^{*} \right) - \dmpmotiondurationprop \angularvelocity\right) + \quatforcingterm + \quatcouplingterm
\end{align*}
Previously, we perform parameter identification to obtain the parameter vector $\dmpparamset$ of the forcing term $\forcingterm$ of DMP from demonstrations of \textit{nominal} behaviors. In this section, we describe how do we learn the parameter 
of the coupling term $\couplingterm$ from \textit{corrected} behavior demonstrations.
\changemarker{
}
This \changemarker{step} opens up several questions such as: What should the input to the feedback model be? How do we extract the training data for the feedback from human demonstrations? What representation should we endow the feedback model with, such that it is able to take into account a high-dimensional sensory input, but also maintain convergence properties of the dynamic motion primitive? 

In this section, we address these questions and present a general learning-from-demonstration framework for the feedback model. We start with the feedback model input specification, then explain how to extract training data, followed by the description of the neural network structure that we use as the feedback model representation, and finally present the loss function for supervised training of the feedback model.
\begin{figure}[ht]
	\centering
    \includegraphics[width=\columnwidth]{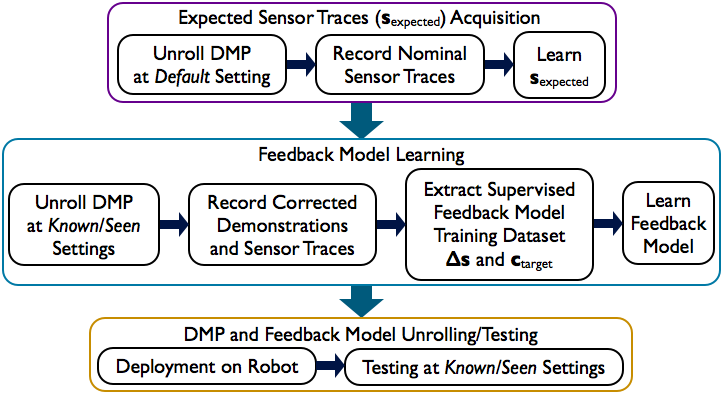}
    \caption{Flow diagram of Phase 2: the initial supervised learning of the behavior feedback models with sensor trace deviation as input. Components of this diagram are explained in section~\ref{ssec:lfb_model_input_spec}, \ref{ssec:lfb_target_ct_extraction_from_demo} and \ref{ssec:lfb_model_representation}. Definition of \textit{Default} and  \textit{Known/Seen} settings can be seen in section~\ref{ssec:env_setting_defn_demo_w_sensor_traces}.}
	\label{fig:FlowDiagramLearnFbTerms_Supervised}
\end{figure}
\subsection{Feedback Model Input Specification}
\label{ssec:lfb_model_input_spec}
In order to perform adaptation, in each time step the feedback model requires some information --regarding the state between the environment and the robot-- related to the task in consideration. This task-related information is obtained via the robot's sensors in the form of the observation of a set of variables that we call as sensor features $\sensorfeatures$, whether they are directly or indirectly related to the task. As mentioned before, in this paper the input of the feedback model is the sensor trace deviations $\sensortracesdeviation = \sensoract - \sensorexp$.

The core idea of Associative Skill Memories (ASMs) \citep{pastor_IROS_2011_ASM, pastor2013dynamic} rests on the insight that similar task executions should yield similar sensory events. Thus, an ASM of a task includes both a movement primitive as well as the expected sensor traces $\sensorexp$ associated with this primitive's execution in its default environment setting.

When a movement primitive is executed under environment variations and/or uncertainties, the online-perceived/actual sensor traces $\sensoract$ tend to deviate from $\sensorexp$. The disparity $\sensoract - \sensorexp = \sensortracesdeviation$ can be used to drive corrections for adapting to the environmental changes causing the deviated sensor traces. This can be written as:
\begin{equation}
    \couplingterm = h(\sensortracesdeviation) = h(\sensoract - \sensorexp)
    \label{eq:sensor_trace_deviations_fb}
\end{equation}
$\couplingterm$ is the degree of adaptation/correction of the behavior, e.g. the coupling term 
$\quatcouplingterm$ in Equation \ref{eq:OriDMPTransformationSystem}.
For this reason, we need to acquire the expected sensor traces $\sensorexp$ beforehand by learning from the sensor traces associated with the nominal behaviors, as shown in the purple block in Figure~\ref{fig:FlowDiagramLearnFbTerms_Supervised}.

To learn the $\sensorexp$ model, we execute the nominal behavior and collect the experienced sensor measurements. Since these measurements are trajectories by nature\footnote{\changemarker{In general, sensor traces can be quite noisy and contain high-frequency components. In the context of this paper, we focus on learning behaviors that depend on low-frequency signals, such that noise can simply be filtered out by a low-pass filter. On the other hand, learning behaviors that depend on high-frequency signals are still challenging and left for future work.}}, we can encode them using DMPs to become $\sensorexp$. This has the advantage that $\sensorexp$ is phase-aligned with the motion primitives' execution.

As a part of preparation for the learning-from-demonstration framework of the feedback model, we collect the dataset of the actual sensor traces $\sensoract$ associated with the adapted/corrected behavior demonstrations, which we call as $\sensortraces_\text{actual, demo}$.
\subsection{Extracting Feedback Model Training data from Demonstrations}
\label{ssec:lfb_target_ct_extraction_from_demo}
Here we assume that the motion primitive's parameter vector $\dmpparamset$ has been learned beforehand from the nominal/baseline behavior demonstrations, following the description in Section \ref{sec:rlfb_background_dmp} and \ref{sec:lfb_extraction_of_movement_primitives_w_auto_demo_segmentation}, and therefore we can already obtain the behavior forcing term $\forcingterm$, following Equation \ref{eq:DMPForcingTerm}.

To perform the learning-from-demonstration of the feedback model, we need to extract the target adaptation level or the target coupling term --which is the target output/regression variable-- from demonstrations data, as follows (in reference to Eq. 
\ref{eq:OriDMPTransformationSystem}):
\begin{align*}
    \targetquatcouplingterm = &-\alpha_{\omega} (\beta_{\omega} 2 \log\left( {\quatevolvinggoalstateposition}_\text{, cd} \circ \quatstateposition_\text{cd}^{*} \right) - \dmpmotiondurationprop_\text{cd} \angularvelocity_\text{cd}) \\
    &+ \dmpmotiondurationprop_\text{cd}^2 \angularacceleration_\text{cd} - \quatforcingterm
    \numberthis
    \label{eq:lfb_OriDMPCouplingTermExtraction}
\end{align*}
where \{$\quatstateposition_\text{cd}, \angularvelocity_\text{cd}, \angularacceleration_\text{cd}$\} is the set of \textit{adapted/corrected} orientation behavior demonstration.
Furthermore, $\dmpmotiondurationprop_\text{cd}$ is the movement duration of each adapted/corrected demonstration, and 
$\alpha_{\omega}$ and $\beta_{\omega}$ are the same constants defined in Section \ref{sec:rlfb_background_dmp}. Due to the specification of $\dmpmotiondurationprop_\text{cd}$ in Eq. 
\ref{eq:lfb_OriDMPCouplingTermExtraction}, the formulation can handle demonstrations with different movement durations/trajectory lengths.

Next, we describe our proposed general learning representations for the feedback model.
\subsection{Feedback Model Representation}
\label{ssec:lfb_model_representation}
Our goal is to acquire a feedback model which predicts the adaptation/correction/coupling term $\couplingterm$ based on sensory information about the environment. In other words, we would like to learn the function $h(.)$, mapping sensory input features $\sensortracesdeviation$ to the coupling term $\couplingterm$, as mentioned in Equation \ref{eq:sensor_trace_deviations_fb}.

\begin{figure}[ht]
	\centering
    \includegraphics[width=0.9\columnwidth]{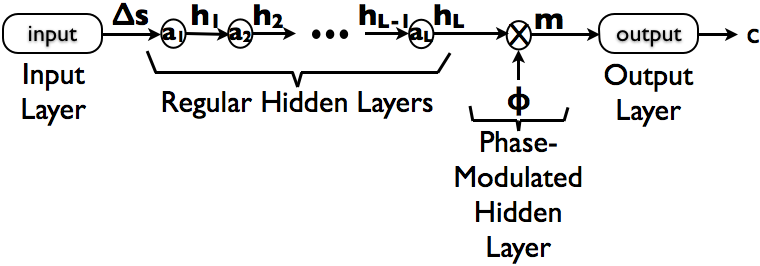}
 	\caption{Phase-modulated neural network ({\pmnn}) with one-dimensional output coupling term $\couplingtermscalar$.}
	\label{fig:lfb_pmnn_special_design_neural_net}
\end{figure}

We continue our previous work \citep{icra2018_learn_tactile_feedback}, using a special neural network structure with embedded post-processing as the feedback model representation. This neural network design is a variant of the radial basis function network (RBFN) \citep{Bishop_1991_RBFNN}, which we call the \textit{phase-modulated neural networks} ({\pmnn}s).
{\pmnn} has an embedded structure that allows encoding of the feedback model's dependency on movement phase as a form of post-processing, whose form is determined automatically from data during learning process.
Moreover, the structure ensures convergence of the adapted motion plan, due to a modulation with the phase velocity.
Diagrammatically, {\pmnn} can be depicted in Figure \ref{fig:lfb_pmnn_special_design_neural_net}.

The {\pmnn} consists of:
\begin{itemize}
	\item \textit{input layer}\\
	    The input is $\sensortracesdeviation = \sensoract - \sensorexp$\changemarker{, a vector of size $N_s$}.
	\item \textit{regular hidden layers}\\
    	The regular hidden layers perform non-linear feature transformations on the high-dimensional inputs.
    	If there are $L$ layers, \changemarker{and there are $N_l$ nodes in the $l$-th layer, then} the output of $l$-th layer is:
    	\begin{equation}
    	\boldsymbol{h}_{l} = \begin{cases}
    	\boldsymbol{a}_{l}\left( \boldsymbol{W}_{\boldsymbol{h}_{l} \sensortracesdeviation} \sensortracesdeviation + \boldsymbol{b}_{\boldsymbol{h}_{l}}  \right) &\text{for $l=1$}\\
    	\boldsymbol{a}_{l}\left( \boldsymbol{W}_{\boldsymbol{h}_{l} \boldsymbol{h}_{l-1}} \boldsymbol{h}_{l-1} + \boldsymbol{b}_{\boldsymbol{h}_{l}}  \right) &\text{for $l=2,...,L$}
    	\end{cases}
    	\end{equation}
    	$\boldsymbol{a}_{l}$ is the activation function of the $l$-th hidden layer, which can be \textrm{tanh}, \textsc{ReLU}, or others. $\boldsymbol{W}_{\boldsymbol{h}_{1} \sensortracesdeviation}$ is the weight matrix \changemarker{of size $N_1 \times N_s$} between the input layer and the first hidden layer. $\boldsymbol{W}_{\boldsymbol{h}_{l} \boldsymbol{h}_{l-1}}$ is the weight matrix \changemarker{of size $N_l \times N_{l-1}$} between the $(l-1)$-th hidden layer and the $l$-th hidden layer. $\boldsymbol{b}_{\boldsymbol{h}_{l}} $ is the bias vector \changemarker{of size $N_l$} at the $l$-th hidden layer.
	\item \textit{final hidden layer with phase kernel modulation}\\
    	This special and final hidden layer takes care of the dependency of the model on the movement phase.
    	The output of this layer is $\boldsymbol{m}$, which is defined as:
    	\begin{equation}
    	\boldsymbol{m} = \normalizeddmpbasisfunc \odot \left( \boldsymbol{W}_{\boldsymbol{m} \boldsymbol{h}_{L}} \boldsymbol{h}_{L} + \boldsymbol{b}_{\boldsymbol{m}}  \right)
    	\label{eq:lfb_pmnn_phase_modulation_layer}
    	\end{equation}
    	where $\odot$ denote element-wise product of vectors\changemarker{, $\boldsymbol{W}_{\boldsymbol{m} \boldsymbol{h}_{L}}$ is a weight matrix of size $N \times N_{l}$, and $\boldsymbol{b}_{\boldsymbol{m}}$ is a bias vector of size $N$}. $\normalizeddmpbasisfunc = \begin{bmatrix} 
    	\normalizeddmpbasisfuncscalar_1 & \normalizeddmpbasisfuncscalar_2 & \hdots & \normalizeddmpbasisfuncscalar_N
        \end{bmatrix}^T$ is the phase kernel modulation vector, and each component $\normalizeddmpbasisfuncscalar_{i}$ is defined as:
    	\begin{equation} 
            \normalizeddmpbasisfuncscalar_{i}\left( \phasevariable, \phasevelocity \right) = \frac{\psi_i \left( \phasevariable \right)}{\sum_{j=1}^N \psi_j \left( \phasevariable \right)} \phasevelocity \qquad i = 1,...,N
          	\label{eq:lfb_pmnn_CouplingTermPhaseModulator}
        \end{equation}
        with phase variable $\phasevariable$ and phase velocity $\phasevelocity$, which comes from the second-order canonical system defined in Equation \ref{eq:2ndOrderCanonicalSystemP1} and \ref{eq:2ndOrderCanonicalSystemP2}. $\psi_i \left( \phasevariable \right)$ is the radial basis function (RBF) as defined in Equation \ref{eq:GaussianBasisFunction}.
        We use $N=25$ phase RBF kernels both in the {\pmnn}s as well as in the DMPs representation. The phase kernel centers have equal spacing in time, and we place these centers in the same way in the DMPs as well as in the {\pmnn}s.
	\item \textit{output layer}\\
    	The output of this layer is the one-dimensional coupling term $\couplingtermscalar$, which is defined as:
    	\begin{equation}
    	    \couplingtermscalar = \boldsymbol{w}_{\couplingtermscalar_m}^{T} \boldsymbol{m}
    	\end{equation}
    	$\boldsymbol{w}_{\couplingtermscalar_m}$ is the weight vector \changemarker{of size $N$}. Please note that there is no bias introduced in the output layer, and hence if $\boldsymbol{m} = \boldsymbol{0}$ --which occurs when the phase velocity $\phasevelocity$ is zero-- then the coupling term $\couplingtermscalar$ is also zero. This ensures that the coupling term is initially zero when the primitive is started. The coupling term will also converge to zero because the phase velocity $\phasevelocity$ is converging to zero. This ensures the convergence of the adapted motion plan.
\end{itemize}
For an $M$-dimensional coupling term, we use $M$ separate {\pmnn}s with the same input vector $\sensortracesdeviation$ and the output of each {\pmnn} corresponds to each dimension of the coupling term. This separation of neural network for each coupling term dimension allows each network to be optimized independently from each other.

\subsection{Supervised Learning of Feedback Models}
\label{ssec:lfb_model_supervised_learning}
The feedback model parameters $\pmnnparamset$ can then be acquired by supervised learning that optimizes the following loss function:
\begin{align*}
    \mathcal{L}_\text{SL} &= \sum_{n=1}^{N_d} \norm{{\targetcouplingterm}_n - h({{\sensortracesdeviation}_\text{demo}}_n; \pmnnparamset)}_2^2 \\
    &= \sum_{n=1}^{N_d} \norm{{\targetcouplingterm}_n - h({\sensortraces_\text{actual, demo}}_n - {\sensorexp}_n; \pmnnparamset)}_2^2 \numberthis
    \label{eq:lfb_supervised_learning_loss_function}
\end{align*}
with $N_d$ is the number of data points available in the adapted/corrected behavior demonstration dataset.
\section{Reinforcement Learning of Feedback Models}
\label{sec:lfb_rl}
In the previous section, we presented an expressive learning representation of feedback models capable of capturing dependency on the movement phase as well as ensuring convergence of the overall behaviors.
Moreover, we presented a method to train the feedback models in a supervised manner, by learning from demonstrations of corrected behaviors.

In this section, we show how to refine the feedback model via a sample-efficient reinforcement learning (RL) algorithm, after the initialization by the supervised learning process. 
Our method extends the previous works on approaches for improving nominal behaviors via RL or Iterative Learning Control \citep{Theodorou_2010_PI2generalized, Theodorou_PI2, Kober_PoWER, Stulp_PI2CMA, Stulp_2012_DMP_RL, Hazara_PI2_DMP_InContactSkills, Gams_TransRob_2014, gams2014learning, Likar_DMPCouplingTermILC, Bristow_ILC, Abu_Dakka2015, Queisser_ILC_DMP_ICDL18} into an RL method for learning feedback models. This extension has two benefits. 
First, in the previous works on the improvement of nominal behaviors, if a new environment situation is encountered, the nominal behavior has to be re-learned. Our method --in contrast-- learns a single adaptive behavior policy which is more general and will be able to tackle multiple different environmental settings via adaptation of the nominal behavior. 
Second, due to our feedback model representation learning, our method maintains its performance on the previous settings, while expanding the range of the adaptive behavior to a new setting. On the other hand, previous methods which refine the nominal behavior will be able to perform well on a new setting, but may not perform well on the previous setting it has seen before.

Our sample-efficient RL algorithm for feedback models performs the optimization in the lower dimensional space, i.e. in the space of the weights of the radial basis functions (RBFs) centered on the phase variable $\phasevariable$ --which is equal in size to the parameter $\dmpparamset$ of the DMP forcing term--, instead of the high-dimensional neural network parameter $\pmnnparamset$ space. This is similar in spirit with the PIGPS algorithm \citep{Chebotar_PIGPS, levine_gps}.
The PIGPS algorithm breaks down the learning of end-to-end policies into three phases: 1) optimization of a low-dimensional policy through  PI\textsuperscript{2} \citep{Theodorou_PI2, Stulp_PI2CMA}, and then 2) rolling out the optimized policies and collecting data tuples of actions and observations; and finally 3) supervised training of the end-to-end vision-to-torque policy based on data collected in phase 2). In our work, we follow a similar process. The main difference is that in our approach, we use the generated trajectory roll-outs to train a \changemarker{local
feedback model around the nominal behavior instead of a global policy}. 
A summary of our approach is presented in Algorithm~\ref{alg:RLFBreactiveMP}, which takes the following as input: 
\begin{itemize}
    \item DMP forcing term parameters $\baselinedmpparamset$ which encodes the \textit{nominal} behavior 
    \item initial feedback model parameters $\pmnnparamset$ 
    \item \textit{corrected} behavior dataset $\correcteddemodataset = \{\quatstateposition_\text{cd}, \angularvelocity_\text{cd}, \angularacceleration_\text{cd}, \sensortracesdeviation\}$ on some known environment settings, which was used to train $\pmnnparamset$ in Eq. \ref{eq:lfb_OriDMPCouplingTermExtraction} and \ref{eq:lfb_supervised_learning_loss_function}
    \item expected sensor trace $\sensorexp$
    \item initial policy exploration covariance $\policyexplorationcovariance$
    \item acceptable cost threshold $\costthreshold$
\end{itemize}
We now describe the phases\footnote{We assume that the environment setting being improved upon is fixed throughout these phases.} of this algorithm in more detail. 
\begin{algorithm}[ht]
	\caption{Reinforcement Learning of Feedback Model for Reactive Motion Planning}
	\label{alg:RLFBreactiveMP}
	\begin{algorithmic}[1]
	    \Function{RLFB}{$\baselinedmpparamset$, $\pmnnparamset$, $\correcteddemodataset$, $\sensorexp$, $\policyexplorationcovariance$, $\costthreshold$}
    		\State $[\trajDMPFBunroll, \_, \cost] \gets \DMPFBunrollfunction(\baselinedmpparamset, \pmnnparamset, \sensorexp)$ \label{aln:initial_closed_loop_behavior_unrolling}
    		\While{$\norm{\cost}_2 > \costthreshold$}
    		    \State $\correcteddmpparamset \gets \DMPtrainfunction(\{\trajDMPFBunroll\})$ \label{aln:conversion_into_loop_behavior}
    		    \State{\# Exploration:}
                		    \For{$k \gets 1$ to $K$}
                		        \State $\indexedperturbedcorrecteddmpparamset \gets \samplingfunction(\normaldistribution(\correcteddmpparamset, \policyexplorationcovariance))$  \label{aln:alg_exploration_sampling}
                		        \State $[\_, \_, \indexedproxycost] \gets \DMPFBunrollfunction\left(\indexedperturbedcorrecteddmpparamset, \zeroparam, \sensorexp\right)$ \label{aln:pi2_sample_unrolling}
                		    \EndFor
                        \State \# $PI^2$ update:
                        \State $[\newcorrecteddmpparamset, \policyexplorationcovariance] \gets \pisquaredcmaupdatefunction(\left\{(\indexedperturbedcorrecteddmpparamset, \indexedproxycost)\right\}_{k=1}^K, \correcteddmpparamset)$ \label{aln:pi2_update}
                        \State $[\newproxytrajDMPFBunroll, \sensoract, \_] \gets \DMPFBunrollfunction\left(\newcorrecteddmpparamset, \zeroparam, \sensorexp\right)$ \label{aln:pi2_improved_param_unrolling}
                    
                    \State $\sensortracesdeviation \gets \sensoract - \sensorexp$ \label{aln:sensortracesdeviation_computation}
                    \State $\additionalcorrecteddemodataset \gets \{\newproxytrajDMPFBunroll, \sensortracesdeviation\}$ \label{aln:fb_model_additional_training_data_creation}
            		\State $\fbmodeltrainingdemodataset \gets \correcteddemodataset + \additionalcorrecteddemodataset$
            		\State $\pmnnparamset \gets \DMPFBtrainfunction(\fbmodeltrainingdemodataset, \baselinedmpparamset)$ \label{aln:supervised_learning_fb_model}
            		\State $[\trajDMPFBunroll, \_, \cost] \gets \DMPFBunrollfunction(\baselinedmpparamset, \pmnnparamset, \sensorexp)$ \label{aln:improved_closed_loop_behavior_unrolling}
    		\EndWhile
    		\State \Return $\pmnnparamset$
    	\EndFunction
	\end{algorithmic}
\end{algorithm}
\begin{algorithm}[ht]
	\caption{Path Integral Policy Improvement with Covariance Matrix Adaptation ($PI^2-CMA$) Update Function}
	\label{alg:PI2CMAupdate}
	\begin{algorithmic}[1]
	    \Function{$\pisquaredcmaupdatefunction$}{$\left\{(\sampleindexeddmpparamset, \indexedcost)\right\}_{k=1}^K, \dmpparamset$}
	        \State $T = length(\indexedcost)$
            \For{$t \gets 1$ to $T$}
                \For{$k \gets 1$ to $K$}
                    \State $\costtogo_{k,t} = \sum_{t=1}^{T} \costscalar_{k,t}$
                    \State $\sampleprobability_{k,t} = \frac{e^{- \frac{1}{\lambda} \costtogo_{k,t}}}{\sum_{k=1}^{K} \left[ e^{- \frac{1}{\lambda} \costtogo_{k,t}} \right]}$
                \EndFor
                \State $\timeindexeddmpparamset^\text{new} = \sum_{k=1}^{K} \sampleprobability_{k,t} \sampleindexeddmpparamset$
                \State $\timeindexedpolicyexplorationcovariance^\text{new} = \sum_{k=1}^{K} \sampleprobability_{k,t} (\sampleindexeddmpparamset - \dmpparamset)(\sampleindexeddmpparamset - \dmpparamset)\T$
            \EndFor
            \State {$\dmpparamset^\text{new} = \frac{\sum_{t=1}^{T} (T-t) \timeindexeddmpparamset^\text{new}}{\sum_{l=1}^{T} (T-l)}$}
            \State {$\policyexplorationcovariance^\text{new} = \frac{\sum_{t=1}^{T} (T-t) \timeindexedpolicyexplorationcovariance^\text{new}}{\sum_{l=1}^{T} (T-l)}$}
            \State \Return $\dmpparamset^\text{new}$, $\policyexplorationcovariance^\text{new}$
        \EndFunction
	\end{algorithmic}
\end{algorithm}

\subsection{Phase 1: Evaluation of the Current Adaptive Behavior and Conversion to a Low-Dimensional Policy}
\label{ssec:rlfb_phase1}
Intuitively, the parameters $\pmnnparamset$ need to capture feedback terms for a variety of settings, and thus a high-capacity feedback model representation is required to represent the feedback variations of many settings. 
However, since we focus on improving the feedback term for the current setting only, we can utilize a lower dimensional representation; 
optimization on the low-dimensional representation --instead of on the high-dimensional feedback model parameters $\pmnnparamset$-- helps us to achieve sample-efficiency in our RL approach.
Therefore, our algorithm first converts the high-dimensional policy into a low-dimensional one, which happens in two steps:
First, the algorithm evaluates the current adaptive behavior based on the high-dimensional policy $\pmnnparamset$ \changemarker{--which includes neural network parameters such as the weights and biases coefficients of each neural network layer of PMNN--}, as is done in line \ref{aln:initial_closed_loop_behavior_unrolling} and \ref{aln:improved_closed_loop_behavior_unrolling} of Algorithm~\ref{alg:RLFBreactiveMP}, which results in the roll-out trajectory $\trajDMPFBunroll$ and the cost per time-step $\cost$.
In the second step, the algorithm compresses the observed trajectory $\trajDMPFBunroll$ into another DMP with low-dimensional forcing term parameters $\correcteddmpparamset$, where $\correcteddmpparamset$ is a set of $25$ weights (see line \ref{aln:conversion_into_loop_behavior} of Alg.~\ref{alg:RLFBreactiveMP}).

\subsection{Phase 2: Optimization of the Low-Dimensional Policy}
\label{ssec:rlfb_phase2}
Once the low-dimensional parametrization $\correcteddmpparamset$ is available, the optimization can be done via the PI\textsuperscript{2} algorithm \citep{Theodorou_PI2}. This consists of three steps: policy exploration, policy evaluation, and policy update. In order to do exploration, we model the noisy version of the transformation system from Eq. \ref{eq:OriDMPTransformationSystem} (without the coupling term $\couplingterm$) as:
\begin{align*}
    \dmpmotiondurationprop^2 \angularacceleration = \alpha_{\omega} \left( \beta_{\omega} 2 \log\left( \quatevolvinggoalstateposition \circ \quatstateposition^{*} \right) - \dmpmotiondurationprop \angularvelocity\right) + (\correcteddmpparamset + \pisquaredpolicynoise)\T \normalizeddmpbasisfunc
\end{align*}
with zero-mean multivariate Gaussian noise $\pisquaredpolicynoise \sim \normaldistribution(\zeroparam, \policyexplorationcovariance)$, 
$\normalizeddmpbasisfunc = \begin{bmatrix} 
    \normalizeddmpbasisfuncscalar_1 & \normalizeddmpbasisfuncscalar_2 & \hdots & \normalizeddmpbasisfuncscalar_N
\end{bmatrix}^T$ is the phase kernel vector, and each component $\normalizeddmpbasisfuncscalar_{i}$ is as defined in Eq.~\ref{eq:lfb_pmnn_CouplingTermPhaseModulator}. In the policy exploration step, we sample $K$ policies from the multivariate Gaussian distribution $\normaldistribution(\correcteddmpparamset, \policyexplorationcovariance)$ as done in line \ref{aln:alg_exploration_sampling} of Alg.~\ref{alg:RLFBreactiveMP}. In the policy evaluation step, we roll-out $K$ trajectories --i.e. one trajectory for each sampled policy--, and evaluate their cost as done in line \ref{aln:pi2_sample_unrolling} of Alg.~\ref{alg:RLFBreactiveMP}. In the policy update step, the algorithm performs a weighted combination of the policy based on the cost: the policies with lower costs are prioritized over those with higher costs, as done in Algorithm~\ref{alg:PI2CMAupdate}\footnote{We follow the variant of the PI\textsuperscript{2} algorithm with covariance matrix adaptation \citep{Stulp_PI2CMA}.}.

\subsection{Phase 3: Rolling Out the Improved Low-Dimensional Policy}
Once the low-dimensional policy is improved, now the algorithm needs to transfer the improvement to the high-dimensional feedback model policy $\pmnnparamset$. 
This transfer is done by rolling out the improved low-dimensional policy $\newcorrecteddmpparamset$ on the real system (line  \ref{aln:pi2_improved_param_unrolling} of Alg.~\ref{alg:RLFBreactiveMP}). This rollout generates the trajectory $\newproxytrajDMPFBunroll$ --which consists of the trajectory of orientation $\quatstateposition$, angular velocity $\angularvelocity$, and angular acceleration $\angularacceleration$-- and the corresponding sensor traces deviation $\sensortracesdeviation$ (line  \ref{aln:sensortracesdeviation_computation} of Alg.~\ref{alg:RLFBreactiveMP}), which are both part of the additional feedback model training data $\additionalcorrecteddemodataset$ (line \ref{aln:fb_model_additional_training_data_creation} of Alg.~\ref{alg:RLFBreactiveMP}). 

\subsection{Phase 4: Supervised Learning of the Feedback Model}
Finally, the algorithm does \emph{supervised} training of the feedback model $\pmnnparamset$ on the combined dataset $\correcteddemodataset + \additionalcorrecteddemodataset$, as is done in line \ref{aln:supervised_learning_fb_model} of Alg.~\ref{alg:RLFBreactiveMP}, following our approach in section~\ref{sec:lfb_model}. Please note that initially --before the RL algorithm was performed--, $\pmnnparamset$ is trained in supervised manner only on the dataset $\correcteddemodataset$, which is the dataset of corrected behaviors on several known environment settings. The additional training data $\additionalcorrecteddemodataset$ will improve the performance of the feedback model $\pmnnparamset$ on the new environment setting where the RL algorithm is performed on.

We repeat these phases until the norm of the cost $\norm{\cost}_2$ converges to either on or below the threshold $\costthreshold$.

\section{System Overview and Experimental Setup}
\begin{figure}[ht]
  \centering
  \includegraphics[width=\columnwidth]{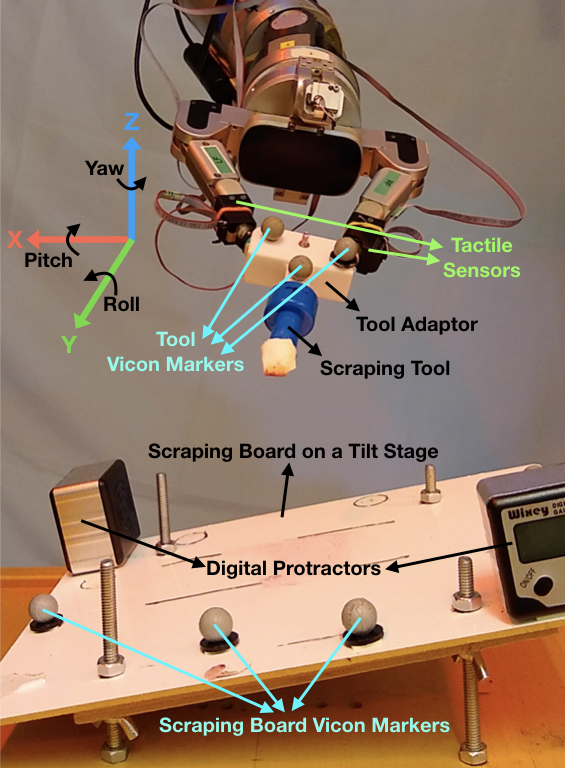}
  \caption{Experimental setup of the scraping task.}
  \label{fig:lfb_pmnn_scraping_setup}
\end{figure}
This work is focused on learning to correct tactile-driven manipulation with tools. As shown in Figure~\ref{fig:lfb_pmnn_scraping_setup}, our experimental scenario involves a demonstrator teaching our robot to perform a scraping task, utilizing a hand-held tool to scrape the surface of a dry-erase board that may be tilted due to a tilt stage, similar to our previous work \citep{icra2018_learn_tactile_feedback}. The system is taught this skill at a default tilt angle, and needs to adapt its behavior when the board is tilted away from that default angle such that it can still scrape the board effectively with the tool. A few important points to note in our experiment:
\begin{itemize}
    \item The system is driven \textit{only} by tactile sensing to inform the adaptation --neither vision nor motion capture system plays a role in driving the adaptation--.
    \item A Vicon motion capture system is used as an external automatic scoring system to measure the performance of the scraping task. The performance in terms of cost for reinforcement learning  is defined in section \ref{sssec:rlfb_cost_defn}.
\end{itemize}
One of the main challenges is that the tactile sensors interact indirectly with the board, i.e. through the tool adapter and the scraping tool via a non-rigid contact, and the robot does not explicitly encode the tool kinematics model. This makes hand-designing a feedback gain matrix for contact control difficult. Next, we explain the experimental setup and some lessons learned from the experiments.

\subsection{Hardware}
The demonstrations were performed on the right arm and the right hand of our bi-manual robot. The arm is a 7-degrees-of-freedom (DoF) Barrett WAM arm which is also equipped with a 6D force-torque (FT) sensor at the wrist.
The hand is a Barrett hand whose left and right fingers are equipped with biomimetic tactile sensors (BioTacs) \citep{wettels08}. The two BioTac-equipped fingers were setup to perform a pinch grasp on a tool adapter. The tool adapter is a 3D-printed object designed to hold a scraping tool with an 11mm-wide tool-tip.

The dry-erase board was mounted on a tilt stage whose orientation can be adjusted to create static tilts of $\pm 10^\circ$ in roll and/or pitch with respect to the robot global coordinates as shown in Figure \ref{fig:lfb_pmnn_scraping_setup}. Two digital protractors with $0.1^\circ$ resolution (Wixey WR 300 Digital Angle Gauge) were used to measure the tilt angles during the experiment.

A set of Vicon markers were placed on the surface of the scraping board, and another set of Vicon markers were placed on the tool adapter. A Vicon motion capture system tracks both sets of Vicon markers in order to compute the relative orientation of the scraping tool w.r.t. the scraping board to evaluate the cost during the reinforcement learning part of the experiment. The cost is defined in section \ref{sssec:rlfb_cost_defn}.

\subsection{Environmental Settings Definition and Demonstrations with Sensory Traces Association}
\label{ssec:env_setting_defn_demo_w_sensor_traces}
For our robot experiment, we considered 7 different environmental \textit{settings}, and each setting is associated with a specific roll angle of the tilt stage, specifically at $0^\circ$, $2.5^\circ$, $5^\circ$, $6.3^\circ$, $7.5^\circ$, $8.8^\circ$, and $10^\circ$. At each setting, we fixed the pitch angle at $0^\circ$ and maintain the scraping path to be roughly at the same height. Hence, we assume that among the 6D pose action (x-y-z-pitch-roll-yaw), the necessary correction/adaptation is only in the roll-orientation.
Here are some definitions of the environmental settings in our experiment:

\begin{itemize}
    \item The \textit{default} setting is the setting where we expect the system to experience the expected sensor traces $\sensorexp$ when executing the nominal behavior without feedback model. We define the setting with roll angle at $0^\circ$ as the default setting, while the remaining settings become the non-default ones.
    \item The \textit{known/seen} settings are a subset of the non-default settings where we collected the demonstration dataset to initialize the feedback model via supervised training.
    \item The \textit{initially unknown/unseen} setting is a non-default setting --disjoint from the known/seen settings-- where the feedback model will be refined on with RL.
    \item The \textit{unknown/unseen} setting is a non-default setting which is located between the \textit{known/seen} settings and the \textit{initially unknown/unseen} setting where we will evaluate the \changemarker{interpolative} generalization capability of the feedback model after the RL process on the \textit{initially unknown/unseen} setting has been done.
    After the RL phase, the feedback model has been trained on both the \textit{known/seen} settings as well as the \textit{initially unknown/unseen} settings. Since the \textit{unknown/unseen} setting is located in between the previous settings, the feedback model is expected to generalize its performance/behavior to some extent to this new setting.
    Moreover, we would argue that the interpolative generalization capability is crucial, since a badly learned model can still overfit and perform poorly in the interpolation setting.
\end{itemize}

For the demonstrated actions, we recorded the 6D pose trajectory of the right hand end-effector at 300 Hz rate, and along with these demonstrations, we also recorded the multi-dimensional sensory traces associated with this action. The sensory traces \changemarker{$\sensortraces$} are the 38-dimensional tactile signals \changemarker{--which are electric signals based on fluid pressures \footnote{\changemarker{Hence $\sensortracesdeviation$ here can be viewed as the change of pressure signals.}}--} from the left and right BioTacs' electrodes, sampled at 100 Hz.

\subsection{Learning Pipeline Details and Lessons Learned}
DMPs provide kinematic plans to be tracked with a position control scheme. However, for tactile-driven contact manipulation tasks such as the scraping task in this paper, using position control alone is not sufficient. This is because tactile sensors require some degree of force control upon contact, in order to attain consistent tactile signals on repetitions of the same task during the demonstrations as well as during the robot's execution.

Moreover, for initializing the feedback model via supervised learning, we need to collect several demonstrations of corrected behaviors at a few of the known/seen non-default settings as described in section \ref{ssec:env_setting_defn_demo_w_sensor_traces}. While it is possible to perform the corrected demonstrations solely by humans, the sensor traces obtained might be significantly different from the traces obtained during the robot's execution of the motion plan. This is problematic, because during learning and during prediction phases of the feedback terms, the input to the feedback models are different. Hence, instead we try to let the robot execute the nominal plans, and only provide correction by manually adjusting the robot's execution at different settings as necessary.

Therefore, we use the force-torque (FT) sensor in the robot's right wrist for FT control, with two purposes: (1) to maintain tool-tip contact with the board, such that consistent tactile signals are obtained, and (2) to provide compliance, allowing the human demonstrator to perform corrective action demonstration as the robot executes the nominal behavior.

For simplicity, we set the force control set points in our experiment to be constant. We need to set the force control set point carefully: if the downward force (in the z-axis direction) for contact maintenance is too big, the friction will block the robot from being able to execute the corrections as commanded by the feedback model. 
We found that 1 Newton is a reasonable value for the downward force control set point.
Regarding the learning process pipeline as depicted in Fig.~\ref{fig:FlowDiagramLearnFbTerms_Overall} and Fig.~\ref{fig:FlowDiagramLearnFbTerms_Supervised}, here we provide the details of our experiment:
\subsubsection{Nominal movement primitives acquisition}\label{sssec:nominal_prim_acquisition}: While the robot is operating in the gravity-compensation mode and the tilt stage is at $0^\circ$ roll angle (the default setting), the human demonstrator guided the robot's hand to kinesthetically perform a scraping task, which can be divided into three stages, each of which corresponds to a movement primitive:
\begin{enumerate}[(a)]
    \item \textit{primitive 1}: starting from its home position above the board, go down (in the z-axis direction) until the scraping tool made contact with the scraping board's surface (no orientation correction at this stage),
    \item \textit{primitive 2}: correct the tool-tip orientation such that it made a full flat tool-tip contact with the surface,
    \item \textit{primitive 3}: go forward in the y-axis direction while scraping the surface, applying orientation correction as necessary to maintain full flat tool-tip contact with the surface.
\end{enumerate}
For robustness, we learn the above primitives --to represent the nominal behavior-- from multiple demonstrations. In particular we collected $\numbaselinedemos = 11$ human demonstrations of nominal behaviors, and use the semi-automated trajectory alignment and segmentation algorithm as mentioned in section \ref{sec:lfb_extraction_of_movement_primitives_w_auto_demo_segmentation}. We extract the \textit{reference} trajectory segments containing primitives 1 and 3 from the first demonstration, by using Zero Velocity Crossing (ZVC) method \citep{fod2002automated} and local minima search refinement on the filtered velocity signal in the z and y axes\footnote{\changemarker{We assume that each primitive has a dominant movement direction, e.g. on z and y axes for primitive 1 and 3, respectively, in this experiment. If it turns out that the primitive's dominant movement direction is multi-dimensional --i.e. spanning in the x-y-z frame--, then we can simply perform a change of coordinate, e.g. such that in the new coordinate frame, the dominant movement direction is in the z axis. Moreover, we assume that each primitive starts and ends with near-zero velocities on that dominant movement direction, so that the ZVC technique is applicable. To determine the ZVC threshold to be used, we plot the first demonstration's --which is manually segmented-- velocity norm w.r.t. the dominant movement direction, and decide on a rough estimate of the ZVC threshold. A rough estimate of the ZVC threshold is fine since it will be followed with the local minima search refinement, which will take care of the rest of the segmentation.}}, to find the initial guesses of the segmentation points of primitives 1 and 3, respectively. The remaining part -- between the end of primitives 1 and the beginning of primitive 3 -- becomes the primitive 2. Afterwards, we perform the automated alignment and weighted least square segmentation\footnote{\changemarker{The hyperparameter --that is involved in the weighted least square segmentation-- is the constant of the exponential decay of the weight. In our experiment, we use the exponential decay constant at value 5.0 to obtain our results.}} on the remaining demonstrations as outlined in section \ref{sec:lfb_extraction_of_movement_primitives_w_auto_demo_segmentation}. We encode each of these primitives with position and orientation DMPs.

\begin{table}[ht]
\centering
\resizebox{\columnwidth}{!}{%
\begin{tabular}{|c|c|c|c|}
\hline
\multirow{2}{*}{} & \multicolumn{3}{c|}{Force-Torque Control Activation Schedule} \\ \cline{2-4}
                  & Prim. 1 & Prim. 2 & Prim. 3 \\ \hline
Step \ref{it:sensor_exp_acquisition} & -   & z 1 N     & z 1 N         \\ \hline
Step \ref{it:supervised_learning_fb} & -   & z 1 N,      & z 1 N,          \\
 &  & roll 0 Nm & roll 0 Nm         \\ \hline
Step \ref{it:rl_fb} & -   & z 1 N     & z 1 N         \\ \hline
Step \ref{it:cl_test} & -   & z 1 N     & z 1 N         \\ \hline
\end{tabular}
\caption{Force-torque control schedule for steps \ref{it:sensor_exp_acquisition}-\ref{it:cl_test}.}
\label{tab:lfb_pmnn_ft_control_schedule_table}
}
\end{table}

For the following steps (\ref{it:sensor_exp_acquisition}, \ref{it:supervised_learning_fb}, \ref{it:rl_fb}, and \ref{it:cl_test}), Table \ref{tab:lfb_pmnn_ft_control_schedule_table} indicates what force-torque control mode being active at each primitive of these steps. "z 1 N" refers to the 1 Newton downward z-axis proportional-integral (PI) force control, for making sure that consistent tactile signals are obtained at repetitions of the task; this is important for learning and making correction predictions properly. "roll 0 Nm" refers to the roll-orientation PI torque control at 0 Newton-meter, for allowing corrective action demonstration.
\subsubsection{Expected sensor traces acquisition}\label{it:sensor_exp_acquisition}:
Still with the tilt stage at $0^\circ$ roll angle (the default setting), 
we let the robot unrolls the nominal primitives 15 times and record the tactile sensor traces. We encode each dimension of the 38-dimensional sensor traces as $\sensorexp$, using the standard DMP formulation \changemarker{with $N=25$ basis functions per dimension}\footnote{\changemarker{This number of basis functions is expressive enough to represent the sensor traces in our case since we are learning behaviors that depend on low-frequency signals, as mentioned in Section \ref{ssec:lfb_model_input_spec}.}}. Please note that once these sensor traces are encoded as DMPs, the encoded DMP parameters are fixed, i.e. never modified nor optimized during the future RL process.
\subsubsection{Supervised feedback model learning on known/seen settings}\label{it:supervised_learning_fb}:
Now we vary the tilt stage's roll-angle to encode each setting in the known/seen environmental settings. At each setting, we let the robot unroll the nominal behavior. 
Beside the downward force control for contact maintenance, now we also activate the roll-orientation PI torque control at 0 Newton-meter throughout primitives 2 and 3. This allows the human demonstrator to perform the roll-orientation correction demonstration, to maintain a full flat tool-tip contact relative to the now-tilted scraping board. We recorded 15 demonstrations for each setting, from which we extracted the supervised dataset for the feedback model, i.e. the pair of the sensory trace deviation ${\sensortracesdeviation}_\text{demo}$ and the target coupling term $\couplingterm_\text{target}$ as formulated in Eq.~\ref{eq:lfb_OriDMPCouplingTermExtraction} and \ref{eq:lfb_supervised_learning_loss_function}. Afterwards, we learn the feedback models from this dataset with {\pmnn}.
\subsubsection{Reinforcement learning of the feedback model on the initially unknown/unseen setting}\label{it:rl_fb}:
We set the tilt stage's roll-angle to encode the initially unknown/unseen setting. Using the reinforcement learning algorithm outlined in section \ref{sec:lfb_rl}, we refine the feedback model to improve its performance over trials in this setting.
\subsubsection{Adaptive behavior (nominal primitive and feedback model) unrolling/testing on all settings}\label{it:cl_test}:
We test the feedback models on the different settings on the robot:
\begin{itemize}
    \item on known/seen settings, whose corrected demonstrations are present in the initial supervised training dataset,
    \item on the initially unknown/unseen setting --on which reinforcement learning was performed on-- and 
    \item on the unknown/unseen setting, which neither was seen during supervised learning nor during reinforcement learning, for testing the \changemarker{interpolative} generalization capability of the feedback model.
\end{itemize}
For the purpose of our evaluations, we evaluate feedback models only on primitives 2 and 3, for roll-orientation correction. 
In primitive 1, we deem that there is no action correction, because the height of the dry-erase board surface is maintained constant across all settings.

\subsection{Learning Representations Implementation}
We implemented all of our models in TensorFlow \citep{TensorFlowBib}, and use \textit{tanh} as the activation function of the hidden layer nodes. We use the Root Mean Square Propagation (RMSProp) \citep{Tieleman2012} as the gradient descent optimization algorithm and set the \textit{dropout} \citep{Srivastava2014_NN_Dropouts} rate to 0.5 to avoid overfitting.

\subsection{Cost Definition for Reinforcement Learning}
\label{sssec:rlfb_cost_defn}
We define the performance/cost of the scraping task as the norm of angular error between the relative orientation of the scraping tool w.r.t. the scraping board during the current task execution versus the relative orientation during the nominal task execution on the default environment setting. If the relative orientation of the scraping tool w.r.t. the scraping board during the current task execution at time $t$ is denoted as Quaternion $\quatstateposition_{cr,t}$, and the relative orientation of the scraping tool w.r.t. the scraping board during the nominal task execution on the default environment setting at time $t$ is denoted as Quaternion $\quatstateposition_{nr,t}$, then the cost at time $t$ is:
\begin{equation}
    \costscalar_t = \norm{2 \log \left( \quatstateposition_{nr,t} \circ \quatstateposition_{cr,t}^{*} \right)}_2
\end{equation}
The cost vector $\cost$ in Algorithm \ref{alg:RLFBreactiveMP} is defined as:
\begin{equation}
    \cost = 
    \begin{bmatrix}
        \costscalar_1 & \costscalar_2 & \hdots & \costscalar_T
    \end{bmatrix}\T
\end{equation}
These relative orientations are measured by a Vicon motion capture system throughout the execution of the tasks.

The reason of the selection of this form of cost is because in order to do the scraping task successfully, the robot shall maintain the particular relative orientation of the scraping tool w.r.t. the scraping board similar to the nominal demonstrations.
\section{Experiments}
\label{sec:rlfb_exp}
After previously presenting the system overview and experimental setup, here we present the experimental evaluations of each components in our learning feedback models pipeline: the semi-automated extraction of nominal movement primitives from demonstrations, the supervised learning of feedback models, and the reinforcement learning of feedback models.
\subsection{Extraction of Nominal Movement Primitives by Semi-Automated Segmentation of Demonstrations}
\label{ssec:exp_auto_demo_segmentation}
\begin{figure*}[ht]
    \centering
    \begin{subfigure}[b]{0.33\textwidth}
        \centering
        \includegraphics[trim={0.0cm 0.0cm 0.0cm 0.0cm},clip,width=\textwidth]{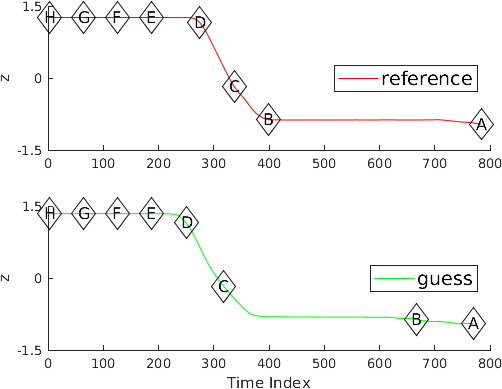}
        \caption{Several (un-weighted) correspondence pairs between the reference segment (top) and a guess segment (bottom).}
        \label{sfig:auto_demo_align_compound_d00898_prim1_unweighted_correspondence_pair}
        \vspace*{0.5cm}
    \end{subfigure}
    \begin{subfigure}[b]{0.33\textwidth}
        \centering
        \includegraphics[trim={0.0cm 0.0cm 0.0cm 0.0cm},clip,width=\textwidth]{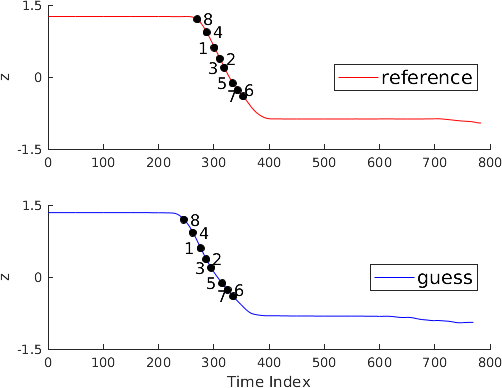}
        \caption{Eight (8) top-ranked weighted correspondence pairs between the reference segment (top) and a guess segment (bottom).}
        \label{sfig:auto_demo_align_compound_d00898_prim1_weighted_correspondence_pair}
        \vspace*{0.5cm}
    \end{subfigure}
    \begin{subfigure}[b]{0.33\textwidth}
        \centering
        \includegraphics[trim={0.0cm 0.0cm 0.0cm 0.0cm},clip,width=\textwidth]{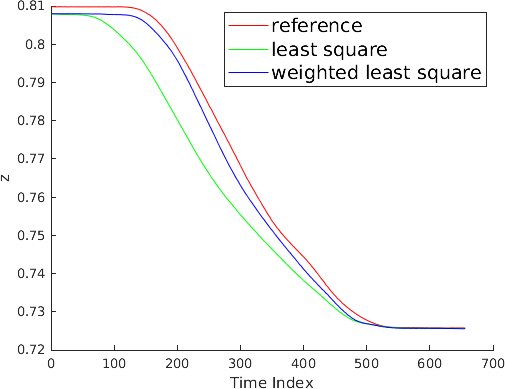}
        \caption{The result of trajectory segmentation refinement (after stretching for comparability) using the regular (un-weighted) least square method versus using the weighted least square method.}
        \label{sfig:auto_demo_align_compound_d00898_prim1_unweighted_vs_weighted_least_square_clipped_aligned}
    \end{subfigure}
    \caption{1 reference segment vs. 1 guess segment: The DTW-computed correspondence matching and refined segmentation results of primitive 1 based on z-axis trajectory, compared between the un-weighted version and the weighted version.}
    \label{fig:auto_demo_align_compound_d00898_prim1}
\end{figure*}
\begin{figure*}[ht]
    \centering
    \begin{subfigure}[b]{0.245\textwidth}
        \centering
        \includegraphics[trim={0.0cm 0.0cm 0.0cm 0.0cm},clip,width=\textwidth]{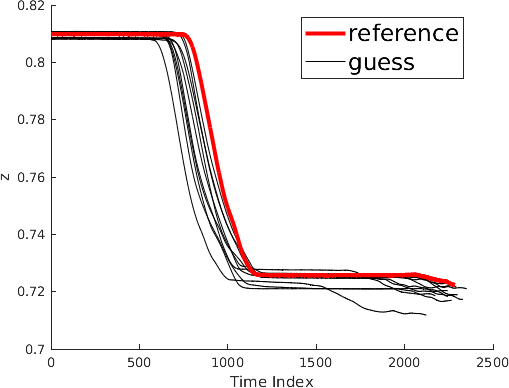}
        \includegraphics[trim={0.0cm 0.0cm 0.0cm 0.0cm},clip,width=\textwidth]{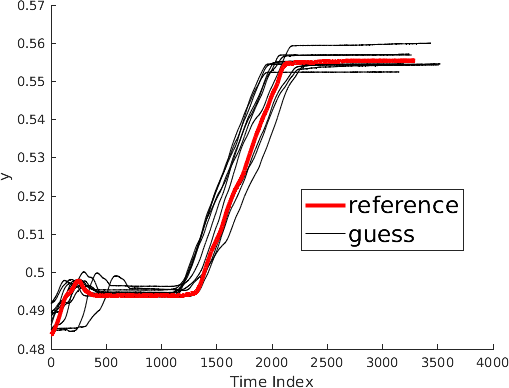}
        \caption{The reference and guess segments before refinement for prim. 1 (top) and prim. 3 (bottom).}
        \label{sfig:auto_demo_align_compound_trajs_prim1_before_refinement}
    \end{subfigure}
    \begin{subfigure}[b]{0.49\textwidth}
        \centering
        \includegraphics[trim={0.0cm 0.0cm 0.0cm 0.0cm},clip,width=0.49\textwidth]{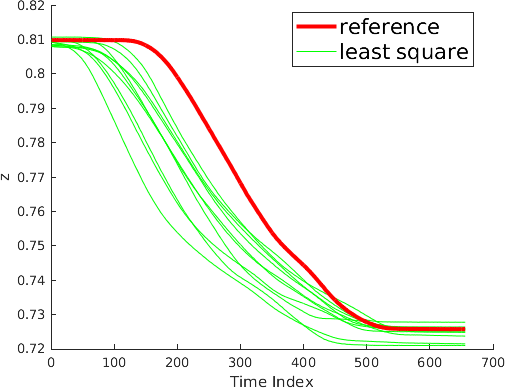}
        \includegraphics[trim={0.0cm 0.0cm 0.0cm 0.0cm},clip,width=0.49\textwidth]{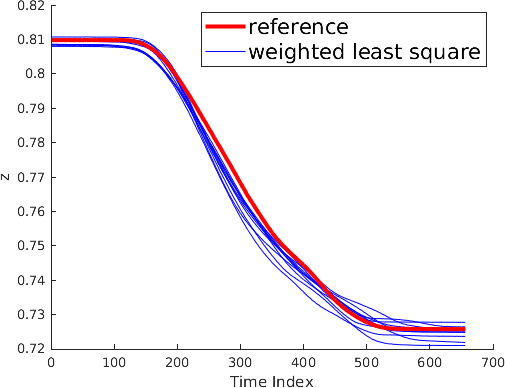}
        \includegraphics[trim={0.0cm 0.0cm 0.0cm 0.0cm},clip,width=0.49\textwidth]{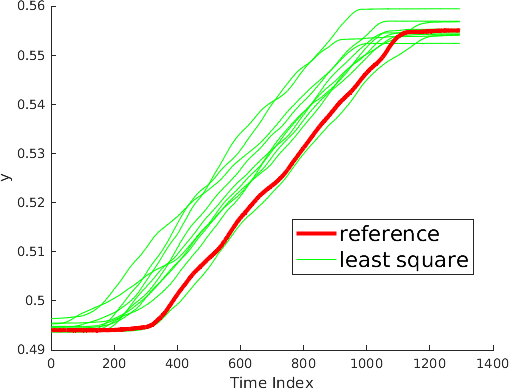}
        \includegraphics[trim={0.0cm 0.0cm 0.0cm 0.0cm},clip,width=0.49\textwidth]{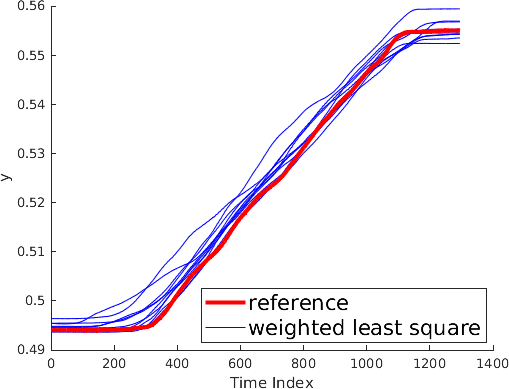}
        \caption{The result of trajectory segmentation refinement (after stretching for comparability) using the regular (un-weighted) least square method (left) versus using the weighted least square method (right) for prim. 1 (top) and prim. 3 (bottom).}
        \label{sfig:auto_demo_align_compound_trajs_prim1_unweighted_vs_weighted_least_square_clipped_aligned}
    \end{subfigure}
    \begin{subfigure}[b]{0.245\textwidth}
        \centering
        \includegraphics[trim={0.0cm 0.0cm 0.0cm 0.0cm},clip,width=\textwidth]{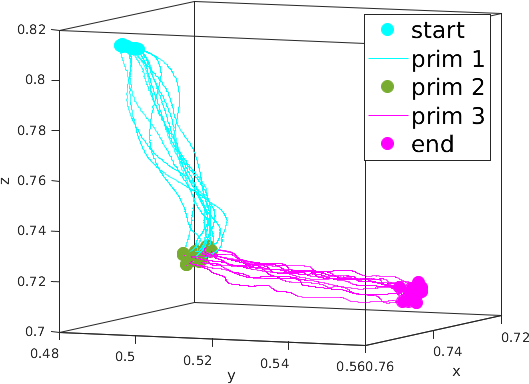}
        \caption{The 3D Cartesian plot of the refined demonstrations' segmentation into primitives.}
        \label{sfig:auto_demo_align_compound_cart_trajs_prims_after_refinement}
    \end{subfigure}
    \caption{1 reference segment vs. 10 guess segments: The refined segmentation results (space vs. time) of primitive 1 (based on z-axis trajectory) and primitive 3 (based on y-axis trajectory), compared between the un-weighted version and the weighted version, as well as the (space-only) 3D Cartesian plot of the segmented primitives.}
    \label{fig:auto_demo_align_compound_trajs_prim1}
\end{figure*}
First, we show the effectiveness of our method for semi-automated segmentation of the nominal behavior demonstrations as outlined in section \ref{sec:lfb_extraction_of_movement_primitives_w_auto_demo_segmentation}.
Our process begins with computing the correspondence matching between the reference segment and each guess segment using the Dynamic Time Warping (DTW) algorithm. For primitive 1, we perform correspondence matching computation on the z-axis trajectory of the end-effector position, while for primitive 3, the matching is done on the y-axis trajectory. This selection is based on the definition of the primitives in section~\ref{sssec:nominal_prim_acquisition}. Once both the segmentation of primitives 1 and 3 are refined, the remaining segment between these two primitives becomes the primitive 2.
\subsubsection{Correspondence Matching Results and Issues}
~\\
Some of the extracted correspondence matches during the segmentation of primitive 1 are shown in Fig.~\ref{sfig:auto_demo_align_compound_d00898_prim1_unweighted_correspondence_pair} and \ref{sfig:auto_demo_align_compound_d00898_prim1_weighted_correspondence_pair}, with numbered pairs as well as alphabet-indexed pairs indicating the matching points.
Among the two possible causes of inaccurate correspondence pairs mentioned in section~\ref{ssec:wls_problem_setup}, the correspondence pair B in Fig.~\ref{sfig:auto_demo_align_compound_d00898_prim1_unweighted_correspondence_pair} is an example of the incorrect matches due to near-zero velocities. 
Including this in the regular least square fashion will result in erroneous alignment and segmentation as can be seen between the red (reference) and green (guess) curves in Figure \ref{sfig:auto_demo_align_compound_d00898_prim1_unweighted_vs_weighted_least_square_clipped_aligned}. 

\subsubsection{Segmentation Results via Weighted Least Square Method}
~\\
To mitigate the negative effect of the erroneous correspondence matches, in section \ref{sec:lfb_extraction_of_movement_primitives_w_auto_demo_segmentation} we proposed to associate each correspondence pair with a weight, and perform a weighted least square method to refine the segmentation. Fig.~\ref{sfig:auto_demo_align_compound_d00898_prim1_weighted_correspondence_pair} shows eight (8) of the top-ranked correspondence pairs based on the weights, meaning that these features will have the most significant influence on the result of the segmentation refinement using the weighted least square method. 
In Fig.~\ref{sfig:auto_demo_align_compound_d00898_prim1_unweighted_vs_weighted_least_square_clipped_aligned}, we show the result of the segmentation refinement using the regular (un-weighted) least square method versus using the weighted least square method, as mentioned in the section \ref{ssec:wls_problem_setup}. The red trajectory is the refined segment of the reference trajectory, while the green and blue trajectories are the refined segments of the initial segmentation guess trajectory using the least square method and the weighted least square method, respectively. Both the green and blue trajectories are displayed in its stretched version, relative to its time duration, so that they become visually comparable to the red trajectory. As can be seen the alignment is much better in the weighted case than the un-weighted one.

Figure~\ref{fig:auto_demo_align_compound_trajs_prim1} shows the segmentation result of all trajectories. 
In Fig.~\ref{sfig:auto_demo_align_compound_trajs_prim1_before_refinement}, we show the reference segment versus all (10) guess segments before the segmentation refinement for primitive 1 (top) and primitive 3 (bottom). 
Figure~\ref{sfig:auto_demo_align_compound_trajs_prim1_unweighted_vs_weighted_least_square_clipped_aligned} shows the result of the segmentation refinement on all guess segments, comparing between using the regular (un-weighted) least square method (left) versus using the weighted least square method (right) on primitive 1 (top) and primitive 3 (bottom); we see that the weighted least square method achieves the superior result as the refined segmentations appear closer together (after the stretching of each refined segments for visual comparability)\footnote{During the supervised training of the nominal primitive parameters $\dmpparamset$, we use the un-stretched version of the segmented trajectories, as the DMP transformation system in Eq.~\ref{eq:OriDMPTransformationSystem} is able to take care of each trajectory time-stretching via the motion duration parameter $\dmpmotiondurationprop$.}. In Fig. \ref{sfig:auto_demo_align_compound_cart_trajs_prims_after_refinement}, we show the nominal behavior demonstrations as well as the primitive segments encoded with colors in the 3D Cartesian space. The primitive 2 only contains orientation motion, thus it looks only like a point in 3D space. In this 3D Cartesian space plot, the segmentation result between the least square and the weighted least square methods are in-distinguishable, and hence we only show the result of the weighted least square method in Fig.~\ref{sfig:auto_demo_align_compound_cart_trajs_prims_after_refinement}. However, as seen in the space-time plots in Fig.~\ref{sfig:auto_demo_align_compound_trajs_prim1_unweighted_vs_weighted_least_square_clipped_aligned}, the weighted least square method achieves superior performance, as the relative time delays between the corresponding demonstration segments --which is an un-modeled phenomenon in learning a DMP from multiple demonstrations-- are minimized.

\subsection{Supervised Learning of Feedback Models}
\label{ssec:supervised_learning_fb_exp}
In accordance with the definition in section \ref{ssec:env_setting_defn_demo_w_sensor_traces}, for all experiments in section \ref{ssec:supervised_learning_fb_exp}, the environmental settings definition are:
\begin{itemize}
    \item The \textit{default} setting is the setting with tilt stage's roll angle at $0^\circ$.
    \item The \textit{known/seen} settings are the settings with tilt stage's roll angle at $2.5^\circ$, $5^\circ$, $7.5^\circ$, and $10^\circ$.
\end{itemize}

To evaluate the performance of the feedback model after supervised training on the demonstration data, first we evaluate the regression and generalization ability of the PMNNs.
Second, we show the superiority of the PMNNs as compared to the regular feed-forward neural networks (FFNNs), for feedback models learning representation. Third, we investigate the importance of learning both the feature representation and the phase dependencies together within the framework of learning feedback models. In comparison to the previous work \citep{icra2018_learn_tactile_feedback}, here we add a comparison of the inclusion of the movement phase information $\phasevariable$ and $\phasevelocity$ as inputs of FFNN versus the inclusion of the movement phase information as radial basis function (RBF) modulation in PMNN.

We use normalized mean squared error (NMSE), i.e. the mean squared prediction error divided by the target coupling term's variance, as our metric. 
To evaluate the learning performance of each model in our experiments, we perform a \textit{leave-one-demonstration-out} test. In this test, we perform $K$ iterations of training and testing, where $K=15$ is the number of demonstrations per setting. At the $k$-th iteration:
\begin{itemize}
    \item The data points of the $k$-th demonstration of all settings are left-out as unseen data for \textit{generalization} testing \emph{across demonstrations}, while the remaining $K-1$ demonstrations' data points\footnote{Each demonstration -- depending on the data collection sampling rate and demonstration duration -- provides hundreds or thousands of data points.} are shuffled randomly and split $85\%$, $7.5\%$, and $7.5\%$ for \textit{training}, \textit{validation}, and \textit{testing}, respectively.
    \item We record the training-validation-testing-generalization NMSE pairs corresponding to the lowest generalization NMSE across learning steps.
\end{itemize}
We report the mean and standard deviation of training-validation-testing-generalization NMSEs across $K$ iterations.
\subsubsection{Regression and Generalization Evaluation of {\pmnn}s}
~\\
The results for primitive 2 and 3, using {\pmnn} structure with one regular hidden layer of 100 nodes, are shown in Table \ref{tab:lfb_pmnn_nmse_table}. The PMNNs achieve good training, validation, testing results, and a reasonable (\emph{across-demonstrations}) generalization results for both primitives. 

\begin{table}[ht]
\centering
\resizebox{\columnwidth}{!}{%
\begin{tabular}{|c|c|c|c|c|}
\hline
\multirow{2}{*}{} & \multicolumn{4}{c|}{Roll-Orientation Coupling Term Learning NMSE} \\ \cline{2-5}
                  & Training & Validation & Testing & Generaliz. \\ \hline
Pr. 2       &0.15$\pm$0.05&0.15$\pm$0.05&0.16$\pm$0.06&0.36$\pm$0.19 \\ \hline
Pr. 3       &0.22$\pm$0.05&0.22$\pm$0.05&0.22$\pm$0.05&0.32$\pm$0.13 \\ \hline
\end{tabular}
\label{table:NMSE}
\caption{NMSE of the roll-orientation coupling term learning with \textit{leave-one-demonstration-out} test, for each primitive (Pr.).}
\label{tab:lfb_pmnn_nmse_table}
}
\end{table}
%
%
\begin{figure}[ht]
    \centering
    \begin{subfigure}{0.5\textwidth}
        \centering
        \includegraphics[trim={3cm 0 4cm 0},clip,width=\textwidth]{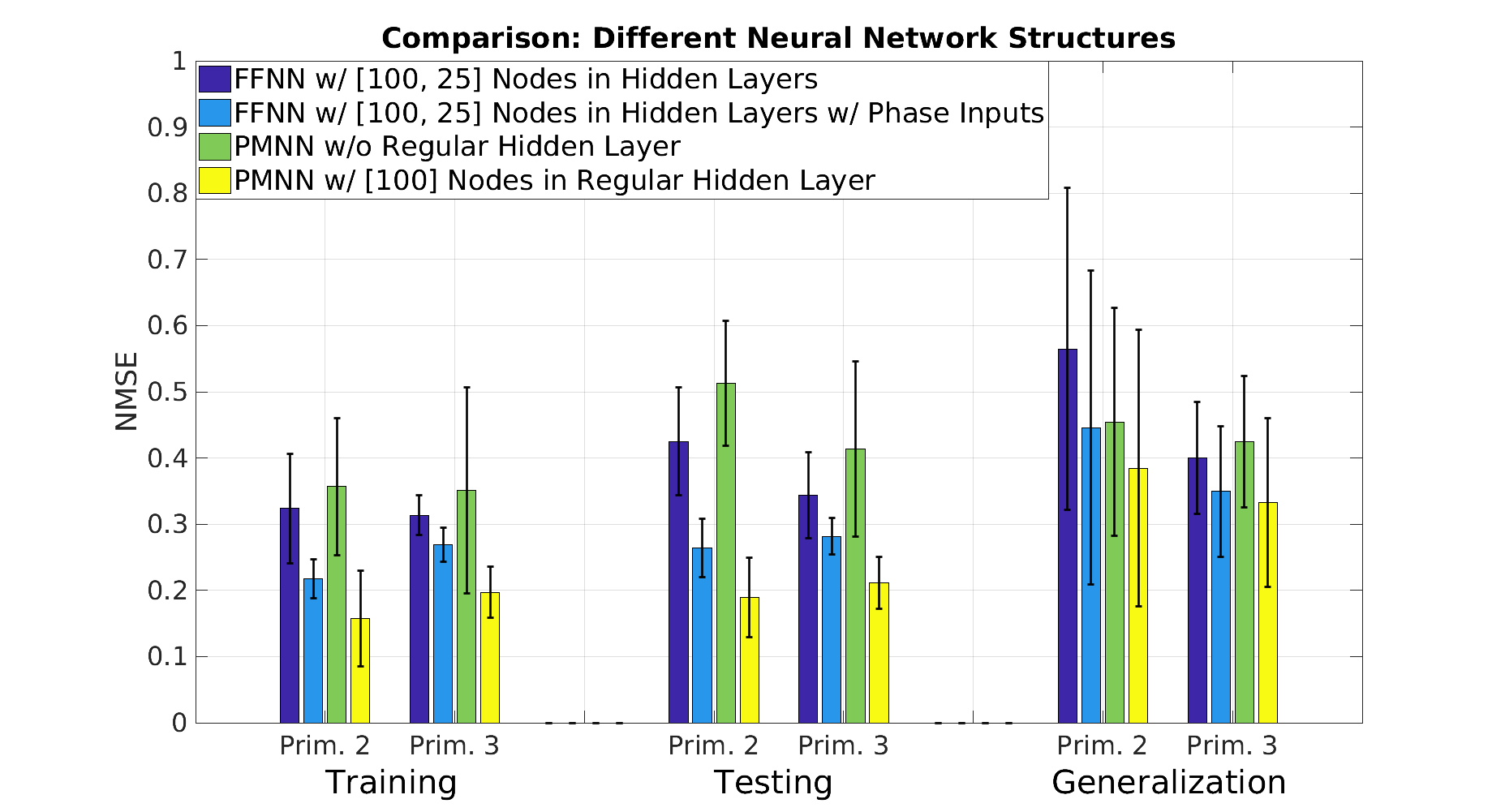}
    \end{subfigure}
    \vspace{0.2cm}
    \begin{subfigure}{0.5\textwidth}
        \centering
        \includegraphics[trim={3cm 0 4cm 0},clip,width=\textwidth]{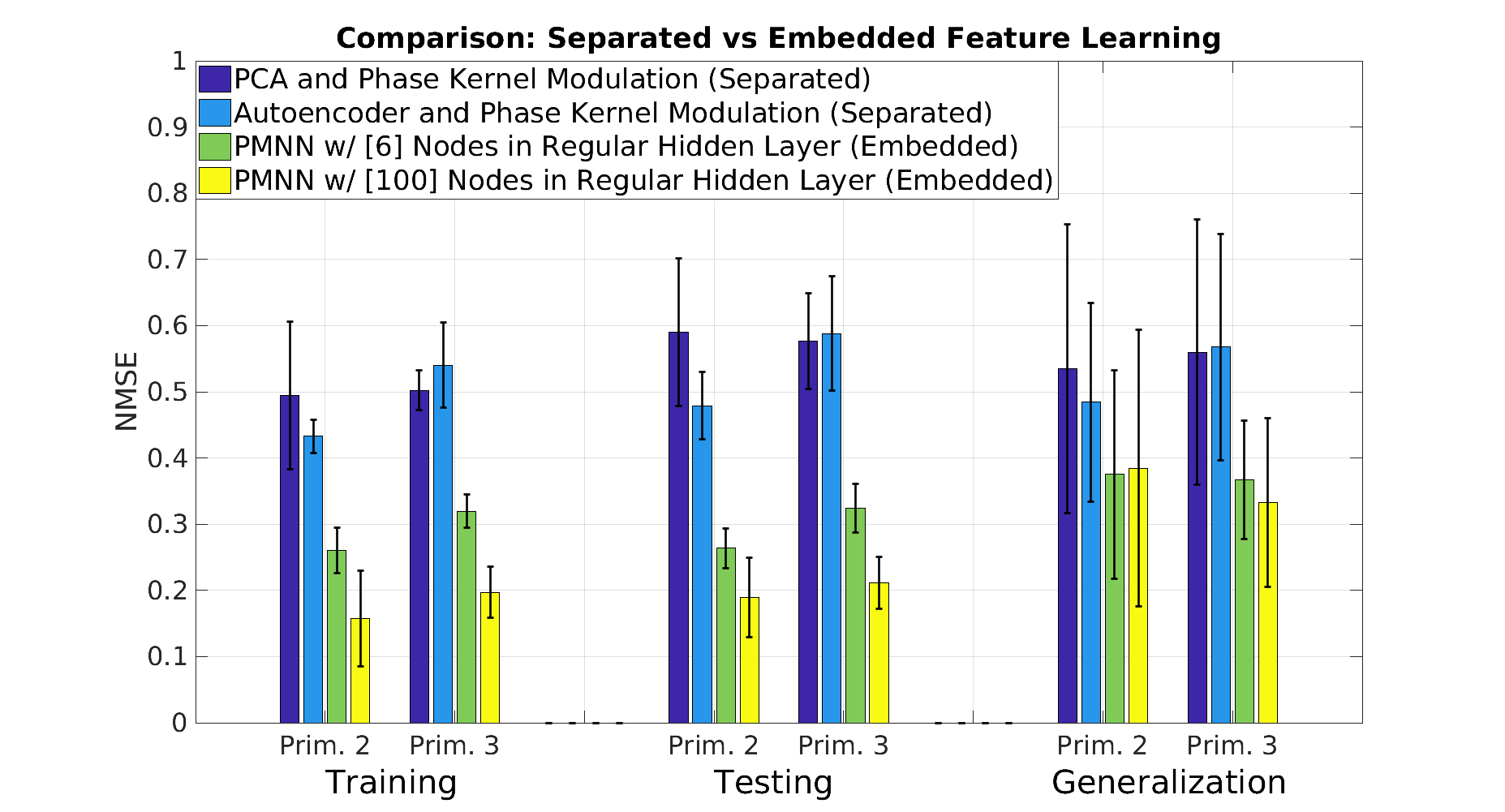}\\
    \end{subfigure}
    \caption{(Top) comparison of regression results on primitives 2 and 3 using different neural network structures; (Bottom) comparison of regression results on primitives 2 and 3 using separated feature learning (PCA or Autoencoder and phase kernel modulation) versus embedded feature learning ({\pmnn})}
    \label{fig:lfb_pmnn_nmse_and_dominant_features}
\end{figure}
\subsubsection{Performance Comparison between FFNN and {\pmnn}}
~\\
We compare the performance between FFNN and {\pmnn}. 
For {\pmnn}, we test two structures: one with no regular hidden layer being used, and the other with one regular hidden layer comprised of 100 nodes. 
For FFNN, we also test two structures, both uses two hidden layers with 100 and 25 nodes each --which is equivalent to {\pmnn} with one regular hidden layer of 100 nodes but de-activating the phase modulation--, but with different inputs: one with only the sensor traces deviation $\sensortracesdeviation$ as 38-dimensional inputs, while the other one with $\sensortracesdeviation$, phase variable $\phasevariable$, and phase velocity $\phasevelocity$ as 40-dimensional inputs. The second FFNN structure is chosen to see the effect of the inclusion of the movement phase information but not as a phase radial basis functions (RBFs) modulation, to compare it with {\pmnn}. The results can be seen in Figure \ref{fig:lfb_pmnn_nmse_and_dominant_features} (Top). It can be seen that {\pmnn} with one regular hidden layer of 100 nodes demonstrated the best performance \changemarker{on average as} compared to the other structures.
The FFNN with additional movement phase information performs significantly better \changemarker{on average} than the FFNN without phase information, which shows that the movement phase information plays an important role in the coupling term prediction. 
However, in general the {\pmnn} with one regular hidden layer of 100 nodes is better \changemarker{on average} than the FFNN, even with the additional phase information.
{\pmnn} with one regular hidden layer of 100 nodes is better \changemarker{on average} than the one without regular hidden layer, most likely because of the richer learned feature representation, without overfitting to the data.

\changemarker{When also taking into account the standard deviation of the performance, the {\pmnn} with one regular hidden layer of 100 nodes is the best in performance in the training and testing cases, but is not very conclusive for the generalization case. However, for {\pmnn} we have the benefit of assurance that the coupling term converges to zero, so the overall behavior will still converge to the goal.}
\subsubsection{Comparison between Separated versus Embedded Feature Representation and Phase-Dependent Learning}
~\\
We also compare the effect of separating versus embedding the feature representation learning with overall parameter optimization under phase modulation. \cite{Chebotar_IROS_2014} used PCA for feature representation learning, which was separated from the phase-dependent parameter optimization using reinforcement learning. On the other hand, {\pmnn} embeds feature learning together with the overall parameter optimization under phase modulation, into an integrated training process.

In this experiment, we used PCA which retained 99\% of the overall data variance, reducing the data dimensionality to 7 and 6 (from originally 38) for primitive 2 and 3, respectively. In addition, we also implemented an autoencoder, a non-linear dimensionality reduction method, as a substitute for PCA in representation learning. For {\pmnn}s, we used two kinds of networks: one with one regular hidden layer of 6 nodes (such that it becomes comparable with the PCA counterpart), and the other with one regular hidden layer of 100 nodes.

Figure \ref{fig:lfb_pmnn_nmse_and_dominant_features} (Bottom) illustrates the superior performance of {\pmnn}s, due to the feature learning performed together with the overall phase-dependent parameters optimization.
\changemarker{Of the two {\pmnn}s in Figure \ref{fig:lfb_pmnn_nmse_and_dominant_features} (Bottom) and the {\pmnn} without regular hidden layer in Figure \ref{fig:lfb_pmnn_nmse_and_dominant_features} (Top), the one with 100 nodes in the regular hidden layer performs the best in training and testing cases even when including the standard deviation into account, and performs mostly the best on average for the generalization case.}

Based on these evaluations, we decided to use {\pmnn}s with one regular hidden layer of 100 nodes and 25 phase-modulated nodes in the final hidden layer for subsequent experiments.
\subsection{Reinforcement Learning of Feedback Models}
\label{ssec:rl_fb_exp}
In accordance with the definition in section \ref{ssec:env_setting_defn_demo_w_sensor_traces}, for all experiments in section \ref{ssec:rl_fb_exp}, the environmental settings definition are:
\begin{itemize}
    \item The \textit{default} setting is the setting with tilt stage's roll angle at $0^\circ$.
    \item The \textit{known/seen} settings are the settings with tilt stage's roll angle at $5^\circ$, $6.3^\circ$, and $7.5^\circ$. These are the settings we performed supervised learning on (as described in section~\ref{sec:lfb_model}), to initialize the feedback model before performing reinforcement learning.
    \item The \textit{initially unknown/unseen} setting is the setting with tilt stage's roll angle at $10^\circ$. This is the novel setting on which we perform the reinforcement learning approach described in section~\ref{sec:lfb_rl}.
    \item The \textit{unknown/unseen} setting is the setting with tilt stage's roll angle at $8.8^\circ$, which neither was seen during the supervised learning nor during the reinforcement learning. We test the final feedback policy after RL on this never-before-seen setting, in order to evaluate the \emph{across-settings} \changemarker{interpolative} generalization capability of the feedback model.
\end{itemize}

Before evaluating our framework for reinforcement learning (RL) refinement of the feedback model, we initialize the feedback model with supervised training dataset collected at the \textit{known/seen} settings. Afterwards, first, we evaluate the learning curve of the feedback model RL refinement on the \textit{initially unknown/unseen} setting.
Second, we compare the feedback models' performance on all non-default settings involved before and after the reinforcement learning, including the \emph{across-settings} \changemarker{interpolative} generalization performance test on the \textit{unknown/unseen} setting.
Finally, we provide snapshots of the real robot execution, comparing the adaptive behavior before and after the RL refinement on the \textit{initially unknown/unseen} setting.

\subsubsection{Quantitative Evaluation of Training with Reinforcement Learning}
~\\
\begin{figure}[ht]
    \centering
    \begin{subfigure}{0.24\textwidth}
        \centering
        \includegraphics[trim={0.0cm 0.0cm 0.0cm 0.0cm},clip,width=\textwidth]{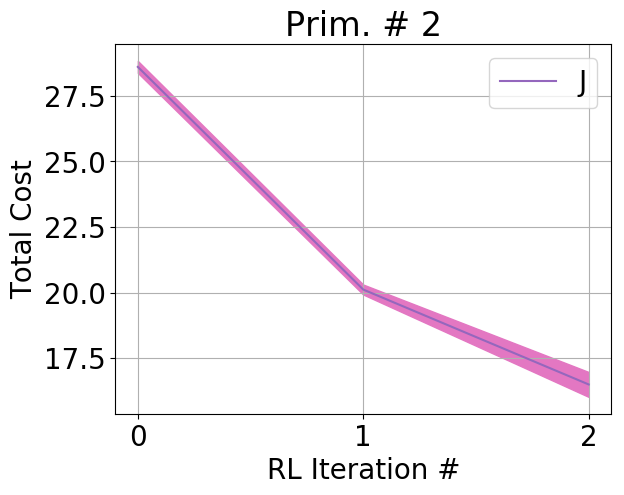}
    \end{subfigure}
    \begin{subfigure}{0.24\textwidth}
        \centering
        \includegraphics[trim={0.0cm 0.0cm 0.0cm 0.0cm},clip,width=\textwidth]{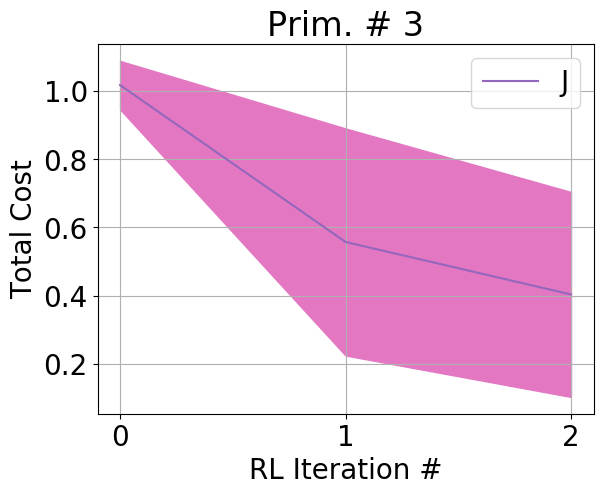}
    \end{subfigure}
    \caption{The learning curves of the RL refinement of the feedback model on the initially unseen setting $10^\circ$ by primitive 2 (left) and 3 (right). The learning curves here shows the mean and standard deviation of the cost over 8 runs on the real robot of the adaptive behavior after the feedback policy update. The cost at iteration \# 0 shows the cost before RL is performed.}
    \label{fig:rl_fb_learning_curves}
\end{figure}

Figure \ref{fig:rl_fb_learning_curves} shows the learning curves of the feedback model refinement by reinforcement learning (RL) on its usage in the adaptive behavior on the initially unseen setting $10^\circ$, on primitive 2 and 3, with the mean and standard deviation computed over 8 runs on the real robot.
The robot first performs RL on the feedback model of the primitive 2, and once its performance converged in primitive 2, the robot performs RL on the feedback model of the primitive 3.

We use $K = 38$ as the number of policy samples taken in the $PI^2-CMA$ algorithm, in Algorithms \ref{alg:RLFBreactiveMP} and \ref{alg:PI2CMAupdate}. An iteration of the while-loop in Algorithm \ref{alg:RLFBreactiveMP} usually takes around 30-45 minutes to complete by the robot (most of the time is taken by the evaluation of the $K$ sampled policies, since the robot needs to unroll each sampled policy in order to evaluate it). The behavior has already converged by the end of the second iteration both on primitive 2 and 3 --in the sense that further iterations do not show significant improvements--, hence we only show the learning curves up to RL iteration \# 2. Hence in total the RL refinement process takes around 2-3 hours to complete on the robot.

This shows the sample efficiency of our proposed reinforcement learning algorithm for the feedback models, since many of the deep reinforcement learning techniques nowadays require millions of samples to complete which is infeasible to be executed on a robot. This is enabled by the fact that we are optimizing the low-dimensional parameters of $\correcteddmpparamset$ --which is of size 25-- via RL as mentioned in section \ref{ssec:rlfb_phase1} and section \ref{ssec:rlfb_phase2} instead of the thousands parameters of {\pmnn}.

\subsubsection{Feedback Models Performance Before versus After Reinforcement Learning and Across-Settings Generalization Performance}
~\\
\begin{figure}[ht]
    \centering
    \begin{subfigure}{0.24\textwidth}
        \centering
        \includegraphics[trim={0.0cm 0.0cm 0.0cm 0.0cm},clip,width=\textwidth]{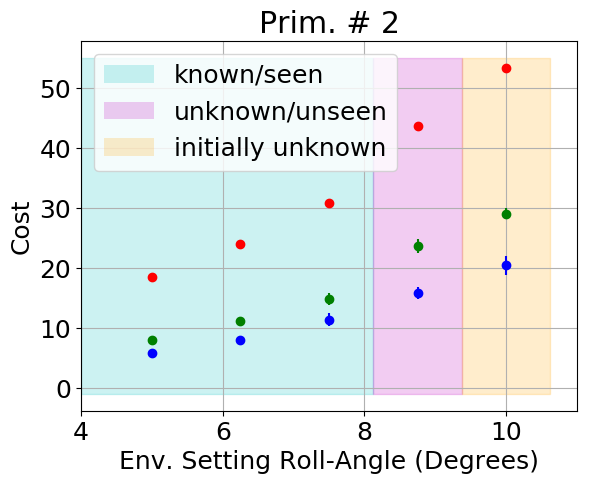}
    \end{subfigure}
    \begin{subfigure}{0.24\textwidth}
        \centering
        \includegraphics[trim={0.0cm 0.0cm 0.0cm 0.0cm},clip,width=\textwidth]{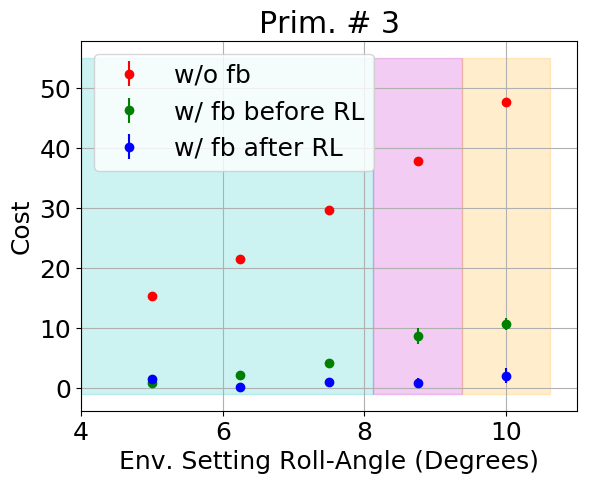}
    \end{subfigure}
    \caption{The performance comparison in terms of accumulated cost on primitive 2 (left) and on primitive 3 (right) between the nominal behavior \emph{without} feedback model (red) vs. the adaptive behavior (including the feedback model) \emph{before} reinforcement learning (RL) of the feedback model (green) vs. the adaptive behavior \emph{after} RL of the feedback model (blue) on all non-default settings. The mean and standard deviation is computed over 8 runs on the real robot.}
    \label{fig:fb_before_vs_after_rl_and_generalization}
\end{figure}

In Figure \ref{fig:fb_before_vs_after_rl_and_generalization}, we compare the performance in terms of the accumulated cost on primitive 2 and primitive 3 at the execution on all non-default settings, between the nominal behavior \emph{without} feedback model (red), the adaptive behavior \emph{before} the reinforcement learning (RL) of the feedback model (green), and the adaptive behavior \emph{after} the RL of the feedback model (blue),  with the mean and standard deviation statistics computed over 8 runs on the real robot. We see that:
\begin{itemize}
    \item The nominal behavior without feedback model (red) always performs worse than the adaptive behaviors with the learned feedback models (green and blue), showing the effectiveness of the learned feedback models.
    \item The performance of the adaptive behavior with the learned feedback model on the initially unseen setting $10^\circ$ improves significantly after the feedback model is refined by RL.
    \item The performance of the adaptive behavior with the learned feedback model on the seen settings $5^\circ$, $6.3^\circ$, and $7.5^\circ$ does not degrade --and even improves-- after the feedback model is refined by RL, as compared to the performance before RL.
    \item The performance of the adaptive behavior with the learned feedback model on the unseen setting $8.8^\circ$ --which was neither seen during the initial supervised learning nor during the reinforcement learning of the feedback model-- is in general in between the performance of its neighboring settings $7.5^\circ$ and $10^\circ$ both before and after RL, which shows some degree of \changemarker{interpolative} generalization across environment settings of the learned feedback model.
\end{itemize}

\begin{figure*}[ht]
    \centering
    \begin{subfigure}{0.24\textwidth}
        \centering
        \includegraphics[width=\textwidth]{before_prim1.png}
        \caption{Before Prim. 1}
    \end{subfigure}
    \begin{subfigure}{0.24\textwidth}
        \centering
        \includegraphics[width=\textwidth]{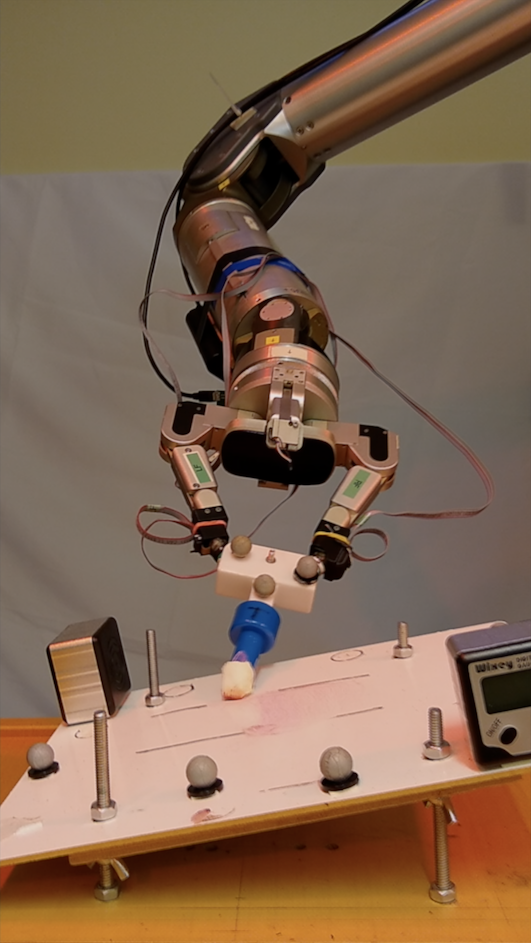}
        \caption{End of Prim. 1}
    \end{subfigure}
	\begin{subfigure}{0.24\textwidth}
        \centering
        \includegraphics[width=\textwidth]{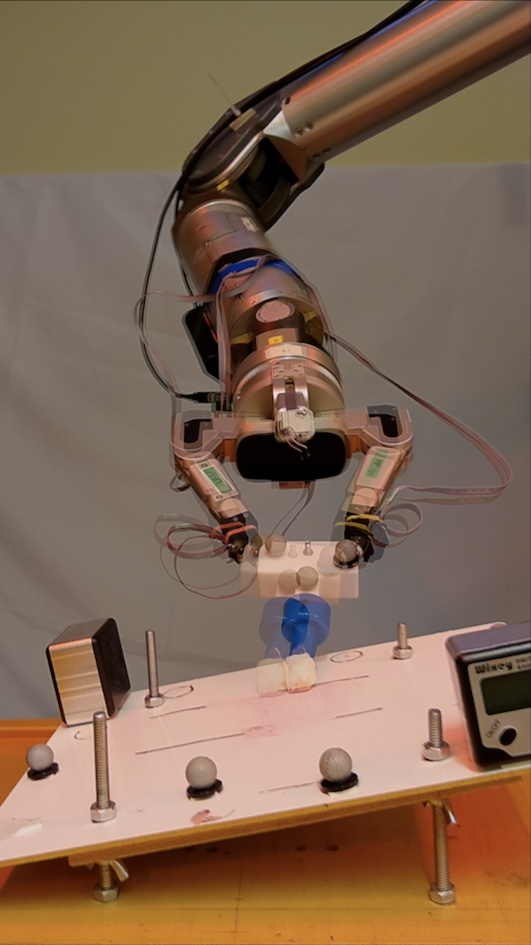}
        \caption{End of Prim. 2}
    \end{subfigure}
	\begin{subfigure}{0.24\textwidth}
        \centering
        \includegraphics[width=\textwidth]{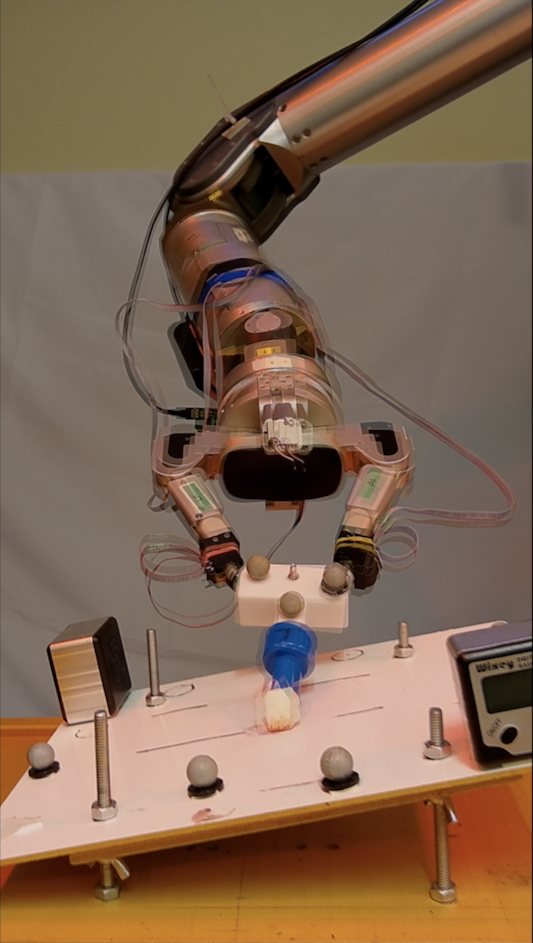}
        \caption{End of Prim. 3}
    \end{subfigure}
    \caption{Snapshots of our experiment on the real robot, comparing the execution of the adaptive behavior (the nominal behavior and the learned feedback model) before RL (soft shadow) versus after RL. After RL, the feedback model applied more correction as compared to the one before RL, qualitatively showing the improvement result by the RL algorithm.}
    \label{fig:rlfb_real_robot_snapshots_before_vs_after_rl}
\end{figure*}
\subsubsection{Qualitative Evaluation of the Real Robot Behavior}
~\\
Figure \ref{fig:rlfb_real_robot_snapshots_before_vs_after_rl} shows the snapshots of our anthropomorphic robot executing the adaptive behavior --that is the nominal behavior and the learned feedback model-- at different stages in the \textit{initially unknown/unseen} setting ($10^\circ$). We compare the behavior before RL (shown as soft shadows) and after RL refinement of the feedback model. As we can see in Fig. \ref{fig:rlfb_real_robot_snapshots_before_vs_after_rl}(c), the feedback model after RL applied more correction than the one before RL, showing the qualitative result of the improvement by the RL algorithm. The quantitative performance improvement due to RL can be seen in Fig. \ref{fig:fb_before_vs_after_rl_and_generalization}.
The pipeline of the RL experiment can be seen in the video \url{https://youtu.be/yu5v-ZXo4-E} or in Extension 1.

\section{Conclusion and Future Work}
\label{sec:rlfb_conclusion}
\changemarker{In conclusion,} we show the effectiveness of our method of learning feedback models for reactive motion planning on a learning tactile feedback testbed, executed on a real robot. In particular, we show that the feedback model can be initialized by supervised learning from demonstrations on several known settings, and then can also be refined further by a sample-efficient reinforcement learning to improve its performance on a setting not seen during the supervised training phase, while retaining its performance on the initial known settings. We also show that the learned feedback model can generalize \changemarker{in the sense of interpolation} to a setting that was neither seen during the supervised training phase nor during the reinforcement learning phase. Furthermore, we detailed the full pipeline of our method, including a weighted least square method for semi-automated segmentation of human demonstrations of nominal behaviors into movement primitives along with its algorithmic analysis and a speed-up possibility.

\changemarker{Regarding future work, one possibility is to extend the experiment in order to evaluate the extrapolative generalization capability of the proposed method. Another possibility is to evaluate the method on a multi-modalities sensing setup, for example combining tactile and vision modalities as inputs. Moreover, so far we have not addressed cases where the demonstrations came from multi-modal distributions, for example in the case of obstacle avoidance, one demonstration goes around one side, while the other demonstration goes around the other side of the obstacle. Evaluating our method on these multi-modal distribution cases is another venue for future work. Finally, learning behaviors that depend on high-frequency signals are still challenging and should be addressed in a future work.}

\begin{funding}
    This research was supported in part by the National Science Foundation grants IIS-1205249, IIS-1017134, EECS-0926052, the Office of Naval Research, and the Okawa Foundation, all of which were issued to the University of Southern California. Moreover, this research was also supported in part by the Max-Planck-Society through funding provided to Giovanni Sutanto, Katharina Rombach, Yevgen Chebotar, Zhe Su, and Stefan Schaal.
\end{funding}

\bibliographystyle{SageH}
\bibliography{references.bib}

\appendix
\section*{APPENDIX}
\section{Quaternion Algebra}
\label{ap:quaternion_algebra} 
Unit quaternion is a hypercomplex number which can be written as a vector $\quatstateposition = 
\begin{bmatrix} 
	r & \boldsymbol{q}^T 
\end{bmatrix}^T$, such that $\|\quatstateposition\| = 1$ with $r$ and $\boldsymbol{q} = \begin{bmatrix} 
	q_1 & q_2 & q_3
\end{bmatrix}^T$ are the real scalar and the vector of three imaginary components of the quaternions, respectively.
For computation with orientation trajectory, several operations needs to be defined as follows:
\begin{itemize}
	\item quaternion composition operation:
        \begin{equation}
        	\quatstateposition_A \circ \quatstateposition_B = 
        \begin{bmatrix} 
        	r_A	& -q_{A1}	& -q_{A2}	& -q_{A3}	\\
        	q_{A1}	& r_A		& -q_{A3	}	& q_{A2}		\\
        	q_{A2}	& q_{A3}		& r_A		& -q_{A1}	\\
        	q_{A3}	& -q_{A2}	& q_{A1}		& r_A
        \end{bmatrix}
        \begin{bmatrix} 
        	r_B	\\
        	q_{B1} 	\\
        	q_{B2}	\\
        	q_{B3}
        \end{bmatrix}
        \label{eq:QuatComposition}
        \end{equation}
	\item quaternion conjugation operation:
        \begin{equation}
        \quatstateposition^{*} = 
        \begin{bmatrix} 
        	r \\
        	-\boldsymbol{q} 
        \end{bmatrix}
        \label{eq:QuatConjugation}
        \end{equation}
    \item \textit{logarithm mapping} ($\log(\cdot)$ operation), which maps an element of $\textit{SO}$(3) to $\textit{so}$(3), is defined as:
        \begin{equation}
        	\log\left( \quatstateposition \right) = \log\left( 
        	\begin{bmatrix} 
        		r \\
        		\boldsymbol{q} 
        	\end{bmatrix}	
        	\right) = 
        	\frac{\arccos{(r)}}{\sin{(\arccos{(r)})}} \boldsymbol{q}
        	\label{eq:LogMapping}
        \end{equation}
    \item \textit{exponential mapping} ($\exp(\cdot)$ operation, the inverse of $\log(\cdot)$ operation), which maps an element of $\textit{so}$(3) to $\textit{SO}$(3), is defined as:
        \begin{equation}
        	\exp\left( \angularvelocity \right) = 
        	    \begin{bmatrix} 
        			\cos{\left( \| \angularvelocity \| \right)} \\
        			\frac{\sin{\left( \| \angularvelocity \| \right)}}{\| \angularvelocity \|} \angularvelocity
        	    \end{bmatrix}
        	\label{eq:ExpMapping}
        \end{equation}
\end{itemize}
\section{Time Complexity Analysis and Quadratic Speed-Up of the Automated Nominal Demonstrations Alignment and Segmentation via Weighted Least Square Dynamic Time Warping}
\label{ap:auto_demo_segmentation_time_complexity_analysis_and_quadratic_speed_up}
The time complexity for DTW algorithm \citep{Sakoe_ASSP_1978} is $\mathcal{O}(MN)$, with $N$ is the length of the reference segment and $M$ is the length of the initial guessed segment, as defined in section~\ref{ssec:lfb_auto_demo_segmentation_point_correspondence_matching_via_dtw}.

For the weighted least square regression:
\begin{itemize}
	\item There are 2 parameters ($\targettimescale$ and $\targettimedelay$) to be estimated, and the total number of correspondence pairs is $K = minimum(M,N)$. Thus at most the dimensionality of $\mathbf{A}$ is $K \times 2$, the dimensionality of $\mathbf{W}$ is $K \times K$, and the dimensionality of $\mathbf{b}$ is $K \times 1$.
	\item The time complexity for matrix-matrix multiplication $\mathbf{A}\T\mathbf{W}$ is $\mathcal{O}(K)$ because $\mathbf{W}$ is a diagonal matrix.
	\item The time complexity for matrix-matrix multiplication $(\mathbf{A}\T\mathbf{W})\mathbf{A}$ is $\mathcal{O}(K)$.
	\item The time complexity for matrix inversion $\left(\mathbf{A}\T\mathbf{W}\mathbf{A}\right)^{-1}$, i.e. inversion of a $2\times2$ matrix is $\mathcal{O}(1)$.
	\item The time complexity for matrix-vector multiplication $(\mathbf{A}\T\mathbf{W})\mathbf{b}$ is $\mathcal{O}(K)$.
	\item The time complexity for matrix-vector multiplication $\left(\mathbf{A}\T\mathbf{W}\mathbf{A}\right)^{-1}\mathbf{A}\T\mathbf{W}\mathbf{b}$ is $\mathcal{O}(1)$.
\end{itemize}
Thus the time complexity of the overall process is $\mathcal{O}(MN)$, i.e. it is dominated by the time complexity of the DTW algorithm.

However, we can gain a significant speed-up of the overall process by down-sampling both the reference and the guessed segments together by a factor of $g$, such that the new length of the reference segment is $N' = \frac{1}{g} N$ and the new length of the guessed segment is $M' = \frac{1}{g}M$. We need that $N' \gg 2$ and $M' \gg 2$, such that the weighted least square computation is still reasonable. The resulting relative time scale will be un-changed, i.e. $\targettimescale' = \targettimescale$, while the relative time delay is related by $\targettimedelay' = \frac{1}{g} \targettimedelay$. The overall time complexity then becomes $\mathcal{O}(M'N') = \mathcal{O}(\frac{1}{g^2} M N)$, i.e. a quadratic speed-up as compared to the original version.

\end{document}